\documentclass[10pt,journal]{IEEEtran}

\usepackage[dvips]{epsfig}
\usepackage{amsmath, amssymb, amsthm}
\usepackage[tight,footnotesize]{subfigure}
\usepackage{multirow}
\usepackage{longtable}
\usepackage{rotating}
\usepackage{dsfont}
\usepackage{color}
\usepackage{hyperref}
\usepackage{algorithmic}
\usepackage{algorithm}
\renewcommand{\algorithmicrequire}{\textbf{Input:}}
\renewcommand{\algorithmicensure}{\textbf{Output:}}

\begin{document}

\title{Kernel Truncated Regression Representation for Robust Subspace Clustering}

\author{Liangli Zhen, Dezhong Peng, Wei Wang, Xin Yao
\thanks{
Liangli Zhen is with the Institute of High Performance Computing, A*STAR, Singapore 138632 (e-mail: llzhen@outlook.com).

Dezhong Peng is with the Machine Intelligence Laboratory, College of Computer Science, Sichuan University, Chengdu 610065, China and the Peng Cheng Laboratory, Shenzhen 518055, China. (pengdz@scu.edu.cn).

Wei Wang is with the Machine Intelligence Laboratory, College of Computer Science, Sichuan University, Chengdu 610065, China and the Chengdu Sobey Digital Technology Co., Ltd., Chengdu 610041, China.

Xin Yao is with the Department of Computer Science and Engineering, Southern University of Science and Technology, Shenzhen 518055, China and the CERCIA, School of Computer Science, University of Birmingham, Birmingham B15 2TT, UK. (e-mail: x.yao@cs.bham.ac.uk).
}
}
\maketitle
\begin{abstract}
Subspace clustering aims to group data points into multiple clusters of which each corresponds to one subspace. Most existing subspace clustering approaches assume that input data lie on linear subspaces. In practice, however, this assumption usually does not hold. To achieve nonlinear subspace clustering, we propose a novel method, called kernel truncated regression representation. Our method consists of the following four steps: 1) projecting the input data into a hidden space, where each data point can be linearly represented by other data points; 2) calculating the linear representation coefficients of the data representations in the hidden space; 3) truncating the trivial coefficients to achieve robustness and block-diagonality; and 4) executing the graph cutting operation on the coefficient matrix by solving a graph Laplacian problem. Our method has the advantages of a closed-form solution and the capacity of clustering data points that lie on nonlinear subspaces. The first advantage makes our method efficient in handling large-scale datasets, and the second one enables the proposed method to conquer the nonlinear subspace clustering challenge. Extensive experiments on six benchmarks demonstrate the effectiveness and the efficiency of the proposed method in comparison with current state-of-the-art approaches.
\end{abstract}


\IEEEpeerreviewmaketitle

\section{Introduction}
Subspace clustering is one of the most popular techniques for data analysis, which has attracted increasing interests of researchers from various areas, such as computer vision, image analysis, and signal processing~\cite{Parsons2004}. With the assumption that high-dimensional data are approximately drawn from a union of low-dimensional subspaces, subspace clustering aims to seek a set of subspaces to fit a given data set and performs clustering based on the identified subspaces~\cite{soltanolkotabi2014robust}.

During the past decades, numerous approaches have been developed for subspace clustering, which can be roughly classified into the following four categories: iterative approaches~\cite{Fan2016}, statistical approaches~\cite{Rao2008}, algebraic approaches~\cite{Vidal2005} and spectral clustering-based approaches~\cite{Von2007, zhen2014locally}. In recent years, spectral clustering-based approaches have attracted more attention and achieved state-of-the-art performance in image clustering and motion segmentation~\cite{zhou2018deep}. The key of this kind of approaches is to find a block-diagonal affinity matrix, where the element of the matrix denotes the similarity between two corresponding data points and the block-diagonal structure means that only the similarity between two intra-cluster data points is nonzero.

To obtain a block-diagonal affinity matrix, some researchers proposed to measure the similarity using the self-expression strategy in spectral clustering methods. Specifically, they represent each data point as a linear combination of the other points in the dataset itself and then use the representation coefficients to build the affinity matrix. The key difference among those methods is the constraint conducted on the representation coefficients and/or the manner of handling noises. For example, sparse subspace clustering (SSC)~\cite{Elhamifar2013} assumes that each data point can be linearly represented by fewest other points and places the $\ell_1$-norm minimisation constraint on the coefficient vectors. Low-rank representation (LRR)~\cite{Liu2013} encourages the coefficient matrix to be low rank for capturing the global structure of the input database. To obtain the low rankness, LRR enforces a nuclear-norm minimisation constraint on the coefficient matrix. Different from SSC and LRR, least squares regression (LSR)~\cite{Lu2013} and truncated regression representation (TRR)~\cite{Peng2016} take $\ell_2$-norm instead of the $\ell_1$-norm and the nuclear-norm to constrain the representation coefficients. The main difference between LSR and TRR is that TRR has a truncated operation on the representation coefficient matrix. Most of these approaches handle the noise by minimising the $\ell_1$-/$\ell_2$-norm of the reconstruction error. To handle the complex noise, He~\textit{et al.}~\cite{He2015} proposed to maximise the correntropy between a given data point and its reconstruction with other points, and Wang~\textit{et al.}~\cite{Wang2015Min} proposed to minimise the entropy of the error between observation signal and its estimation. In addition, Li~\textit{et al.}~\cite{li2015robust} proposed to exploit the intrinsic geometric structure of data and the local and global structural consistencies over labels to learn discriminative features.  Kang~\textit{et al.}~\cite{kang2019robust} proposed to remove the noise and errors from the raw data adaptively using low-rank recovery and robust principal  component analysis to achieve robust graph learning. Furthermore, from the point view of semi-supervised learning, \cite{li2017robust} explores both the labelled and unlabelled to explicitly learn the block-diagonal structure in a nonnegative matrix factorization~(NMF) framework. In recent years, some research works are developed to achieve clustering on the large-scale datasets. For instance, Peng~\textit{et al.}~\cite{peng2013scalable} extended SSC by using the sampling strategy.  Kang~\textit{et al.}~\cite{kang2019large} used a smaller graph to approximate the full graph adaptively by learning from the raw data. These methods have shown promising performance in subspace clustering.

Note that these subspace clustering methods are originally developed to handle the data that are (approximately) drawn from a union of linear subspaces. They may not be able to obtain a satisfactory clustering result when the input data points lie on a set of nonlinear subspaces. In real-world systems, however, most of the collected data are located on nonlinear subspaces~\cite{seung2000manifold, kang2017kernel}. It brings a challenge to linear subspace clustering methods and limits their applications in the real world. Even though there are some deep learning-based clustering approaches~\cite{peng2018stru} are developed to tackle this challenge, they usually need a large amount of data to be available and have high computational complexity, which hinders their applications largely.

\begin{figure}[bpt]
\centering
\includegraphics[width=0.95\linewidth]{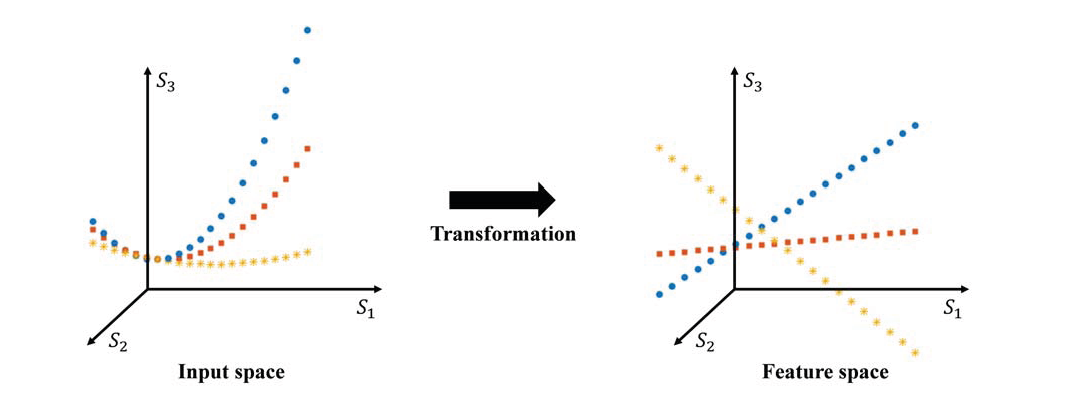}
\caption{The basic idea of our method. By projecting the data into another space with an implicit nonlinear transformation, our method could solve the problem of nonlinear subspace clustering. The left and right plots correspond to the data distributions in the input and hidden space, respectively.}
\label{fig:transformation}
\end{figure}

To cluster the data drawn from a union of nonlinear subspaces, in this paper, we propose a novel robust subspace clustering method, termed kernel truncated regression representation (KTRR). Our basic idea is based on the assumption that there exists a hidden space in which the data can be linearly represented by each other. To illustrate this simple but effective idea, we provide a toy example in Fig.~\ref{fig:transformation}. From the example, we can see that the data points lie on three curves in the input space so that they cannot be linearly represented with each other directly. After a nonlinear transformation, these data points lie on three lines in the hidden space and can be linearly represented by each other. The proposed method consists of the following four steps: 1) projecting the data from the input space into a hidden space in which the mapped data lie on linear subspaces; 2) calculating the global self-expression of the whole data set in the hidden space; 3) eliminating the impact of noise, such as Gaussian noise, by zeroing trivial coefficients; 4) constructing a Laplacian graph using the obtained coefficients and solving a generalised Eigen-decomposition problem to obtain the clustering membership with the algorithm of k-means.

The main contributions and novelty of this work can be summarised as follows:

\begin{itemize}
  \item We propose a new robust subspace clustering method, which can cluster the data points drawn from multiple nonlinear subspaces. By exploiting the kernel technique to transform the input samples into the hidden space, we effectively tracked the challenge that TRR cannot handle the data lie on nonlinear subspaces.

  \item A closed-form solution to KTRR is provided. The solution or the proposed model is a function of the kernel matrix and not directly related to the input data. It makes our method very efficient, especially when handling the problem that has high-dimensional input data.

  \item The proposed KTRR is further extended to handle large-scale datasets. We transform the scalability issue in KTRR as a kind of out-of-sample problem, and solve it with a ``sampling, clustering, and classifying'' strategy.

  \item We apply the KTRR method for several image clustering problems. Extensive experiments are conducted to investigate the effectiveness and efficiency of KTRR. The empirical results show that our method significantly outperforms current state-of-the-art subspace clustering algorithms in terms of accuracy, robustness, and computational cost.
\end{itemize}

The rest of this paper is organised as follows. Section \ref{Sec:2} reviews the related work. Section \ref{Sec:3} is devoted to the formulation of kernel truncated regression and the presentation of the proposed method. Section \ref{Sec:4} reports the experimental results to evaluate the effectiveness and efficiency of the proposed method. Section \ref{Sec:5} concludes the paper.

\textbf{Notations:}
In this paper, unless specified otherwise, \textbf{lower-case bold letters} represent column vectors, \textbf{upper-case bold letters} represent matrices, and the entries of matrices are denoted with subscripts. For instance, $\mathbf{v}$ is a column vector, $v_i$ is its $i$th entry. $\mathbf{M}$ is a matrix, $m_{ij}$ is the entry in the $i$th row, $j$th column, and $\mathbf{m}_j$ denotes the $j$th column of $\mathbf{M}$. Moreover,  $\mathbf{M}^T$ represents the transpose of $\mathbf{M}$, $\mathbf{M}^{-1}$ denotes the inverse matrix of $\mathbf{M}$, and $\mathbf{I}$ stands for the identity matrix. Table~\ref{tab1} summarises some notations used throughout the paper.

\begin{table}[!ht]
\caption{Some notations used in this paper}
\label{tab1}
\center
\begin{center}
\begin{tabular}{ll}
\hline
Notation & Definition\\
\hline
$m$ & dimensionality of input data points\\

$n$ & number of input data points\\

$L$ & number of underlying subspaces\\

$\mathbf{X} \in \mathbb{R}^{m \times n}$  & input data matrix\\

$\mathbf{x}_i \in \mathbb{R}^{m}$  & vector of the $i$th column of $\mathbf{X}$\\

$\mathbf{X}_i \in \mathbb{R}^{m \times n}$  & dictionary matrix for the data point $\mathbf{x}_i$\\

$\kappa(\mathbf{x}_i, \mathbf{x}_j)$  & kernel function\\

$\mathbf{K} \in \mathbb{R}^{n \times n}$  & kernel matrix of the input data matrix\\

$\phi$: $\mathbb{R}^m \rightarrow \mathcal{H}$  & mapping function from the input space to the kernel space\\

$\phi(\mathbf{x}_i)$ & representation of $\mathbf{x}_i $ in the kernel space\\

$\mathbf{c}_i \in \mathbb{R}^{n}$  & representation coefficient vector for $\phi(\mathbf{x}_i)$\\

$\mathbf{C} \in \mathbb{R}^{n \times n}$  & linear representation coefficient matrix\\

$\mathbf{W} \in \mathbb{R}^{n \times n}$  & similarity matrix of input data matrix\\

$\mathbf{L} \in \mathbb{R}^{n \times n}$  & normalized Laplacian matrix\\

\hline
\end{tabular}
\end{center}
\end{table}

\section{Related Work}
\label{Sec:2}

In the past decades, a large number of spectral clustering-based methods have been proposed to achieve subspace clustering in many applications,  such as image clustering, motion segmentation and gene expression analysis~\cite{Elhamifar2013}. The key of these methods is to obtain a block-diagonal similarity matrix whose nonzero elements are only located on the connections of the points from the same subspace. There are two common strategies to compute the similarity matrix, \textit{i.e.}, pairwise distance-based strategy and linear representation-based strategy. Pairwise distance-based strategy computes the similarity between two points according to their pairwise relationship. For example, the original spectral clustering method adopts the Euclidean distance with Heat Kernel to calculate the similarity as
\begin{equation}
\label{eq2.0}
s(\mathbf{x}_i, \mathbf{x}_j) = e^{-\frac{\|\mathbf{x}_i - \mathbf{x}_j\|^2}{2\sigma^2}},
\end{equation}
where $s(\mathbf{x}_i, \mathbf{x}_j)$ denotes the similarity between the data points $\mathbf{x}_i$  and $\mathbf{x}_j$,  and the parameter $\sigma$ controls the width of the neighborhoods.

Alternatively, linear representation-based approaches assume that each data point can be represented by a linear combination of some other points from the intra-subspace~\cite{Elhamifar2013, zhen2017under, zhen2020obj}. Based on this assumption, the linear representation coefficient is used as a measurement of similarity. The linear representation-based approaches have achieved promising performance in subspace clustering~\cite{Elhamifar2013, Liu2013,Peng2016,Liu2011} since they encode the global structure of the data set into the representation coefficient matrix.

By given a data matrix $\mathbf{X} = [\mathbf{x}_1, \mathbf{x}_2, \dots \mathbf{x}_n] \in R^{m \times n}$, the representation-based methods linearly represent $\mathbf{X}$ and obtain the coefficient matrix $\mathbf{C}\in R^{n\times n}$ in a self-expression manner by solving

\begin{equation}
\label{eq2.1} \min \Re(\mathbf{C}) \hspace{3mm} \textup{s.t.}
\hspace{3mm} \mathbf{X} =
\mathbf{X}\mathbf{C}, \text{diag}(\mathbf{C}) = \mathbf{0},
\end{equation}
where $\text{diag}(\mathbf{C}) = \mathbf{0}$ avoids the trivial solution that uses the data point to represent itself.
 $\Re(\mathbf{C})$ denotes the adopted prior structured regularisation on $\mathbf{C}$, and the major difference among most existing subspace clustering methods is the choice of $\Re(\mathbf{C})$. For example, SSC~\cite{Elhamifar2013} enforces the sparsity on the column vectors of $\mathbf{C}$ by adopting $\ell_1$-norm via $\Re(\mathbf{C})= \sum_{i=1}^n \|\mathbf{c}_i\|_{1}$, LRR~\cite{Liu2013} obtains low rankness by using nuclear-norm with $\Re(\mathbf{C})=\|\mathbf{C}\|_{\ast}$. To further achieve robustness, (\ref{eq2.1}) is extended as follows:
\begin{equation}
\label{eq2.2} \min \Re(\mathbf{C}) + \wp(\mathbf{E}) \hspace{3mm} \textup{s.t.}
\hspace{3mm} \mathbf{X} =
\mathbf{X}\mathbf{C} + \mathbf{E}, \text{diag}(\mathbf{C}) = \mathbf{0},
\end{equation}
where $\mathbf{E}$ stands for the errors induced by the noise and corruption, $\wp(\mathbf{E})$ measures the impact of the errors. Generally, $\wp(\cdot)=\|\mathbf{E}\|_{F}$ and  $\wp(\cdot)=\|\mathbf{E}\|_{1}$ are used to handle the Gaussian noise and the Laplacian noise, respectively, and $\|\cdot\|_{F}$ denotes the Frobenius-norm of a matrix.

Due to the assumption of linear reconstruction, those methods failed to achieve nonlinear subspaces clustering. To address this challenging issue, some research have conducted few attempts~\cite{patel2014kernel, Xiao2015}. However, these methods are computationally inefficient since they need to solve $\ell_1$- or nuclear-norm minimisation problems. Some deep learning-based methods~\cite{zhou2018deep,Ji2017deep,peng2018stru} are also developed to learn nonlinear relationships. They usually need a large amount of data available to explore the nonlinear relationships of the data and cost a lot on computational resources. To address these issues, we propose a new nonlinear subspace clustering method by integrating the kernel technique into linear representation.

\section{The proposed subspace clustering method}
\label{Sec:3}
This section presents the details of our proposed method. Firstly, we introduce the formulation and the optimisation procedure of KTRR. Then, we illustrate how to use the representation coefficient matrix of KTRR to achieve robust subspace clustering. Next, the computational complexity of the proposed method is analysed. At last, an extension of KTRR is provided for handling large-scale datasets.

\subsection{Kernel Truncated Regression Representation}

For a given data set $\{\mathbf{x}_i\}_{i=1}^n$, where $\mathbf{x}_i \in \mathbb{R}^m$, we define a matrix $\mathbf{X} = [\mathbf{x}_1, \mathbf{x}_2, \dots \mathbf{x}_n]$. Let $\phi$: $\mathbb{R}^m \rightarrow \mathcal{H}$ be a nonlinear mapping which transforms the input data into a kernel space $\mathcal{H}$, and $\phi(\mathbf{X}_i) = [\phi(\mathbf{x}_1), \dots, \phi(\mathbf{x}_{i-1}), \mathbf{0}, \phi(\mathbf{x}_{i+1}), \dots, \phi(\mathbf{x}_n)]$. After mapping $\mathbf{X}$ into a kernel space, the corresponding $\{\phi(\mathbf{x}_i)\}_{i=1}^n$ is generally believed lying in multiple linear subspaces if an appropriate transforming function is selected~\cite{patel2014kernel,Xiao2015}. Based on this basic idea, we propose to formulate the objective function of our KTRR as follows:
\begin{equation}
\label{equ:3.2} \mathop{\mathrm{min}}_{\mathbf{c}_i}
\hspace{1mm}\frac{1}{2}\|\phi(\mathbf{x}_i) -
\phi(\mathbf{X}_i)\mathbf{c}_i\|_2^2+\frac{\lambda}{2}\|\mathbf{c}_i\|_2^2,
\end{equation}
where the first term is the reconstruction error in the kernel space, the second term serves as an $\ell_2$-norm regularization, and $\lambda$ is a positive real number, which controls the strength of the $\ell_2$-norm regularization term.

For each transformed data representation $\phi(\mathbf{x}_{i})$, solving the optimisation problem (\ref{equ:3.2}), it gives that

\begin{equation}
\label{equ:3.22}
\mathbf{c}_{i}^* = \left(\phi(\mathbf{X}_{i})^{T}\phi(\mathbf{X}_{i}) + \lambda \mathbf{I}\right)^{-1} \phi(\mathbf{X}_{i})^{T} \phi(\mathbf{x}_{i}).
\end{equation}
One can find that the solution in (\ref{equ:3.22}) does not require $\phi(\mathbf{x}_i)$ to be explicitly computed. It only needs the dot products of the images of the data in the hidden space. The dot products can sometimes be calculated more efficiently as a direct function of the input data points, without explicitly performing the mapping $\phi$. In other words, the computation of the images in the hidden space can be by-passed. A function that performs this direct computation is a kernel function. For some choices of a kernel $\kappa(\mathbf{x}_i,\mathbf{x}_j)$: $\mathbb{R}^m \times \mathbb{R}^m \mapsto \mathbb{R}$, \cite{Shawe2004} has shown that $\kappa$ can obtain the dot product in the kernel space $\mathcal{H}$ induced by the mapping $\phi$.

For example, for a set of input data points $\{\mathbf{x}_i\}_{i=1}^n$ lie in a two-dimensional space, we project these points into another space with the mapping function
\begin{equation}
\label{equ:3.23}
\phi: \mathbf{x}_{i} = (x_{1i}, x_{2i})^T \mapsto \phi(\mathbf{x}_{i}) = (x_{1i}^2, x_{2i}^2, \sqrt{2}x_{1i}x_{2i})^T.
\end{equation}
The mapping function projects the data points from a two-dimensional space to a three-dimensional space where the dot product of two images can be computed as
\begin{equation}
\label{equ:3.24}
\begin{split}
\phi(\mathbf{x}_{i})^T\phi(\mathbf{x}_{j}) & = (x_{1i}^2, x_{2i}^2, \sqrt{2}x_{1i}x_{2i})(x_{1j}^2, x_{2j}^2, \sqrt{2}x_{1j}x_{2j})^T\\
& = x_{1i}^2x_{1j}^2 + x_{2i}^2x_{2j}^2 + 2x_{1i}x_{2i}x_{1j}x_{2j}\\
& = (x_{1i}x_{1j} + x_{2i}x_{2j})^2 = (\mathbf{x}_{i}^T\mathbf{x}_{j})^2.
\end{split}
\end{equation}
Therefore, the function $\kappa(\mathbf{x}_i,\mathbf{x}_j) = (\mathbf{x}_i^T\mathbf{x}_j)^2$ is the kernel which can be used to compute the dot products of the images corresponding to the above mapping function $\phi$ in (\ref{equ:3.23}) without explicitly calculating the coordinates of their images. Furthermore, the same kernel computes the dot product corresponding to the following mapping function
\begin{equation}
\label{equ:3.23}
\phi: \mathbf{x}_{i} = (x_{1i}, x_{2i})^T \mapsto \phi(\mathbf{x}_{i}) = (x_{1i}^2, x_{2i}^2, x_{1i}x_{2i}, x_{2i}x_{1i})^T,
\end{equation}
which demonstrates that the same kernel function may corresponding to several different mapping functions~\cite{Shawe2004}. In return, the mapping function is uniquely corresponding to one kernel function. It helps us select a suitable kernel function for the data being processed. Also, using the kernel technique to compute the dot product of images has lower computational complexity by comparing with computing the dot product with the explicitly mapping process. At last, there is no need to change the formulation of our method in (\ref{equ:3.2}) to accommodate the particular choice of kernel function. We can select any suitable kernel for the data set being considered.

It is notable that it still requires $O(n^4 + mn^2)$ to obtain the solution in (\ref{equ:3.22}) for the problem of $n$ data points with dimensionality of $m$. To solve the problem in (\ref{equ:3.2}) more efficiently, we firstly rewrite it as

\begin{equation}
\label{eq3.20}
\min_{\mathbf{c}_i} \frac{1}{2}\|\phi(\mathbf{x}_i) - \phi(\mathbf{X})\mathbf{c}_i\|_2^2 + \frac{\lambda}{2} \|\mathbf{c}_i\|_2^2,
\hspace{3mm} \textup{s.t.} \hspace{3mm} \mathbf{e}_i^T\mathbf{c}_{i}=0,
\end{equation}
where $\phi(\mathbf{X}) = [\phi(\mathbf{x}_1), \phi(\mathbf{x}_2), \dots \phi(\mathbf{x}_n)]$, $\mathbf{e}_i$ is a column vector with all zero elements except the $i$th element is one, and the constraint $\mathbf{e}_i^T\mathbf{c}_{i}=0$ eliminates the trivial solution of representing a transformed point by itself.

Using the Lagrangian method, we obtain that

\begin{equation}
\label{eq3.3}
L(\mathbf{c}_i) = \frac{1}{2}\|\phi(\mathbf{x}_i) - \phi(\mathbf{X})\mathbf{c}_i\|_2^2 + \frac{\lambda}{2} \|\mathbf{c}_i\|_2^2 + \theta \mathbf{e}_i^T\mathbf{c}_{i},
\end{equation}
where $\theta$ is the Lagrangian multiplier. Clearly,

\begin{equation}
\label{eq3.4}
\frac{\partial L(\mathbf{c}_i)}{\partial \mathbf{c}_i} = (\phi(\mathbf{X})^T\phi(\mathbf{X}) + \lambda \mathbf{I})\mathbf{c}_i - \phi(\mathbf{X})^T\phi(\mathbf{x}_i) + \theta\mathbf{e}_i.
\end{equation}

Let $\frac{\partial L(\mathbf{c}_i)}{\partial \mathbf{c}_i} = 0$, we have

\begin{equation}
\label{eq3.5}
\mathbf{c}_i^* = (\phi(\mathbf{X})^T\phi(\mathbf{X}) + \lambda \mathbf{I})^{-1} (\phi(\mathbf{X})^T\phi(\mathbf{x}_i) -\theta\mathbf{e}_i).
\end{equation}

Multiplying $\mathbf{e}_i^T$ on both sides of (\ref{eq3.5}), and since $\mathbf{e}_i^T\mathbf{c}_i = 0$, it holds that

\begin{equation}
\label{eq3.6}
\theta = \frac{\mathbf{e}_i^T(\phi(\mathbf{X})^T\phi(\mathbf{X}) + \lambda \mathbf{I})^{-1} \phi(\mathbf{X})^T\phi(\mathbf{x}_i)}{\mathbf{e}_i^T(\phi(\mathbf{X})^T\phi(\mathbf{X}) + \lambda \mathbf{I})^{-1}\mathbf{e}_i}.
\end{equation}

Substituting (\ref{eq3.6}) into (\ref{eq3.5}), the optimal solution is given as

\begin{equation}
\label{eq3.7}
\mathbf{c}_i^* = \mathbf{q}_i - \mathbf{P} \frac{\mathbf{e}_i^T\mathbf{q}_i\mathbf{e}_i}{\mathbf{e}_i^T\mathbf{P}\mathbf{e}_i},
\end{equation}
where $\mathbf{q}_i = \mathbf{P}(\phi(\mathbf{X})^T\phi(\mathbf{x}_i))$, and $\mathbf{P} = (\phi(\mathbf{X})^T\phi(\mathbf{X}) + \lambda \mathbf{I})^{-1}$.

We can combine all the dot products as a matrix $\mathbf{K} \in \mathbb{R}^{N \times N}$ whose elements are calculated as

\begin{equation}
\label{eq3.8}
K_{ij} = \phi(\mathbf{x}_i)^T \phi(\mathbf{x}_j) = [\phi(\mathbf{X})^T\phi(\mathbf{X})]_{ij} = \kappa(\mathbf{x}_i, \mathbf{x}_j),
\end{equation}
where $\phi(\mathbf{X}) = [\phi(\mathbf{x}_1), \phi(\mathbf{x}_2), \dots \phi(\mathbf{x}_n)]$. The matrix $\mathbf{K}$ is the kernel matrix, which is a symmetric and positive semidefinite matrix. Accordingly, (\ref{eq3.7}) can be rewritten as

\begin{equation}
\label{eq3.9}
\mathbf{c}_i^* = \mathbf{v}_i - \mathbf{U} \frac{\mathbf{e}_i^T\mathbf{v}_i\mathbf{e}_i}{\mathbf{e}_i^T\mathbf{U}\mathbf{e}_i},
\end{equation}
where $\mathbf{U} = (\mathbf{K} + \lambda \mathbf{I})^{-1}$, $\mathbf{v}_i = \mathbf{U}\mathbf{k}_i$, and $\mathbf{k}_i$ is the $i$th column vector of $\mathbf{K}$.

Note that only one pseudo-inverse operation is needed for solving the representation problems of all data points. The computational complexity of calculating the optimal solutions in (\ref{eq3.9}) has decreased to $O(n^3 +mn^2)$ for $n$ data points with $m$ dimensions.

It has been proved that, under certain condition, the coefficients over intra-subspace data points are larger than those over inter-subspace data points~\cite{Peng2016}. After representing the data set by the kernel matrix via (\ref{eq3.9}), we handle the errors by performing a hard thresholding $\mathcal{T}_\eta(\cdot)$ over $\mathbf{c}_i^*$, where $\mathcal{T}_\eta(\cdot)$ keeps $\eta$ largest entries in $\mathbf{c}_i$ and sets other entries as zeros, \textit{i.e.},
\begin{equation}
\label{eq3.11}
\mathcal{T}_\eta(\mathbf{c}_i^*) = [\mathcal{T}_\eta(c_{1i}^*), \mathcal{T}_\eta(c_{2i}^*), \dots, \mathcal{T}_\eta(c_{ni}^*)]^T
\end{equation}
and
\begin{equation}
\label{eq3.12}
\mathcal{T}_\eta(c_{ji}^*) = \begin{cases} c_{ji}^*, & \mbox{if }~~c_{ji}^* \in \Omega_i\mbox{;}\\ 0, & \mbox{otherwise,} \end{cases}
\end{equation}
where $ \Omega_i$ consists of the $\eta$ largest elements of $\mathbf{c}_i^*$.
Typically, the optimal $\eta$ equals to the dimensionality of corresponding kernel subspace, which can be estimated by subspace learning approaches~\cite{peng2017automatic}. In this manner, it avoids to formulate the impact of the noises into the optimisation problem explicitly and does not need prior knowledge about the errors.

\subsection{KTRR for Robust Subspace Clustering}
In this section, we present the method to achieve subspace clustering by incorporating KTRR into the spectral clustering framework~\cite{Ng2002}.

For a given data set $\mathbf{X}$, which consists of $n$ data points in $\mathbb{R}^m$, we assume that these points lie on a union of $L$ low-dimensional nonlinear subspaces. We propose to transform the data points into a hidden space, in which the images of these data points can be linearly represented by the images of the data points from the intra-subspace. From (\ref{eq3.7}), we find that the calculation of the representation coefficients does not require the transforming function in an explicit form, but the dot products of the images are needed. We can induce a kernel function to calculate these dot products and obtain the representation coefficients via (\ref{eq3.9}).

Moreover, the existence of the noises in the input dataset leads to some error connections among the data points from different subspaces. We propose to remove these errors through hard thresholding on each column vector of the coefficient matrix $\mathbf{C}^*$ via (\ref{eq3.11}).

These representation coefficients can be seen as the similarities among the input data points. The similarity between two intra-subspace data points is large, and the similarity between two inter-subspace data points is zero or is approximately equal to zero. Therefore, we can build a similarity matrix $\mathbf{W}$ based on the obtained coefficient matrix $\mathbf{C}^*$ as
\begin{equation}
\label{eq3.13}
\mathbf{W} = |(\mathbf{C}^*)^T| + |\mathbf{C}^*|.
\end{equation}
The matrix of $\mathbf{W}$ is symmetric and is suitable for integrating into the spectral clustering framework.

Then, we compute the normalized Laplacian matrix by following \cite{Ng2002}:
\begin{equation}
\label{eq3.14}
\mathbf{L} = \mathbf{I} - \mathbf{D}^{-\frac{1}{2}}\mathbf{W}\mathbf{D}^{-\frac{1}{2}},
\end{equation}
where $\mathbf{D}$ is a diagonal matrix with $d_{ii} = \sum_{j = 1}^n w_{ij}$. $\mathbf{L}$ is positive semi-definite and has an eigenvalue equals zero with the eigenvector $\mathbf{D}^{\frac{1}{2}} \mathbf{1}$~\cite{Von2007}, where $\mathbf{1} = [1, \dots, 1]^T \in \mathbb{R}^n$.

Next, we calculate the first $L$ eigenvectors $\mathbf{y}_1, \mathbf{y}_2, \dots, \mathbf{y}_L$ of $\mathbf{L}$, which corresponding to its first $L$ smallest nonzero eigenvalues, and construct a matrix $\mathbf{Y} = [\mathbf{y}_1, \mathbf{y}_2, \dots, \mathbf{y}_L] \in \mathbb{R}^{n \times L}$.

Finally, we apply the k-means clustering algorithm to the matrix $\mathbf{Y}$, by treating each row vector of $\mathbf{Y}$ as a data point. In this way, we can cluster the data into $L$ groups and obtain the clustering membership. The proposed subspace clustering algorithm is summarised in Algorithm $1$.

\begin{algorithm}[!ht]
\label{algorithm1}
\renewcommand{\algorithmicrequire}{\textbf{Input:}}
\renewcommand{\algorithmicensure}{\textbf{Output:}}
\caption{Learning kernel truncated regression representation for robust subspace clustering}

    \begin{algorithmic}[1]
    \REQUIRE
    A given data set $\mathbf{X}\in \mathbb{R}^{m \times n}$, the tradeoff parameter $\lambda$, the parameter $\eta$, and the number of subspaces $L$.

    \ENSURE
    The clustering labels of the input data points.

    \STATE Calculate the kernel matrix $\mathbf{K}$ and the matrix $\mathbf{U}$ in (\ref{eq3.9}) and store them.

    \STATE For each point $\mathbf{x}_i\in \mathbb{R}^{m}$, calculate its linear representation coefficients in the kernel space $\mathbf{c}_i^* \in \mathbb{R}^{n}$ via (\ref{eq3.9}).

    \STATE Remove trivial coefficients from $\mathbf{c}_i^*$ by performing hard thresholding $\mathcal{T}_\eta(\mathbf{c}_i^*)$, \textit{i.e.}, keeping $\eta$ largest entries in $\mathbf{c}_i^*$ and zeroing all other elements.

    \STATE Construct a symmetric similarity matrix via (\ref{eq3.13}).

    \STATE Calculate the normalised Laplacian matrix $\mathbf{L}$ via (\ref{eq3.14}).

    \STATE Compute the eigenvector matrix $\mathbf{Y} \in \mathbb{R}^{n \times L}$ that consists of the first $L$ normalised eigenvectors of $\mathbf{L}$, corresponding to its $L$ smallest nonzero eigenvalues.
    \STATE Apply the k-means algorithm to cluster the rows of $\mathbf{Y}$ into $L$ groups and obtain the clustering membership.

    \end{algorithmic}
\end{algorithm}

\subsection{Computational Complexity Analysis}
Given a data matrix $\mathbf{X} \in \mathbb{R}^{m\times n}$, KTRR takes $O(mn^2)$ to compute the kernel matrix $\mathbf{K}$. Then it takes $O(n^3)$ to obtain the matrix $\mathbf{U}$, and $O(mn^2)$ to calculate all the solutions in (\ref{eq3.9}) with the matrices $\mathbf{U}$ and $\mathbf{K}$. Finally, it requires $O(\eta\text{log}\eta)$ to find $\eta$ largest coefficients in each column of the representation matrix $\mathbf{C}^*$. Putting these steps together, we obtain the computational complexity of KTRR as $O(mn^2+n^3)$. This computational complexity is the same as that of TRR, and is considerably less than that of KSSC $(O(mn^2 + tn^3))$\cite{patel2014kernel}, KLRR$(O(t(r_{\mathbf{X}}+r)n^2))$\cite{Xiao2015}, where $t$ denotes the total number of iterations for the corresponding algorithm, $r_{\mathbf{X}}$ is the rank of $\mathbf{X}$, and $r$ is the rank for partial SVD at each iteration of KLRR.

\subsection{Handling Large-Scale Data Sets with KTRR}
From the above computational complexity analysis, we can see that even KTRR is much faster than many existing subspace clustering methods, \textit{e.g.}, KLRR, KSSC, LRR, SSC. However, it is still unable to handle large-scale datasets. Inspired by the strategy in \cite{Peng2015b}, we extend KTRR to handle large-scale datasets with the following three steps: 1) sampling, 2) clustering, and 3) classification.

In the first step, it assumes that the sampled data subset and the whole data set are independent and identically distributed (\textit{i.e.}, i.i.d.) so that the out-of-sample data could be represented by the sampled data. This assumption is general on which most of machine learning algorithms are based. In this paper, we adopt a uniform random sampling approach to sample a subset of data from the input data set. Then, we use the proposed KTRR to obtain the clustering membership of the sampled data. In the third step, we train a classifier with the sampled data and the corresponding labels and classify the out-of-sample data with the trained classifier.

For the classification of out-of-sample data, the adopted algorithm should be capable of handling data that lie on nonlinear subspaces. We propose to take a fully-connected three-layer ($10$ hidden units) feed-forward neural network to achieve the classification task since it is easy to train and is potentially able to handle data which lie on nonlinear subspaces. Some other deep neural networks can also be used in this framework. A potential problem of adopting a deep neural network is that when the dimension of input data is very high, a small number of in-sample data may not be sufficient for learning the parameters of the neural network. The procedure of the extended KTRR (EKTRR) is summarised in Algorithm~2.

\begin{algorithm}[!ht]
\label{algorithm2}
\renewcommand{\algorithmicrequire}{\textbf{Input:}}
\renewcommand{\algorithmicensure}{\textbf{Output:}}
\caption{Clustering large-scale datasets with EKTRR}

    \begin{algorithmic}[1]
    \REQUIRE
    A given data set $\mathbf{X}\in \mathbb{R}^{m \times n}$, the tradeoff parameter $\lambda$, thresholding parameter $\eta$, the number of sampled data points $\delta$, and the number of subspaces $L$.

    \ENSURE
    The clustering labels of the input data points.

    \STATE Randomly select $\delta$ data columns from $\mathbf{X}$ as in-sample data matrix $\mathbf{A}$. The remaining samples are denoted as out-of-sample data $\mathbf{Z} = (\mathbf{z}_1, \mathbf{z}_2, \dots, \mathbf{z}_{n-\delta})$.

    \STATE Perform KTRR (Algorithm 1) on $\mathbf{A}$ to obtain the cluster membership of $\mathbf{A}$.

    \STATE Train a feed-forward neural network with $\mathbf{A}$ and the corresponding labels.

    \STATE Classify each out-of-sample data point in $\mathbf{Z}$ and obtain the cluster membership of $\mathbf{Z}$.

    \end{algorithmic}
\end{algorithm}

\section{Experimental Study}
\label{Sec:4}
In this section, we experimentally evaluate the performance of the proposed method. We consider the results in terms of three aspects: 1) accuracy, 2) robustness, and 3) computational cost. Robustness is evaluated by conducting experiments using samples with two different types of corruptions, \textit{i.e.}, Gaussian noises and random pixel corruption.

\subsection{Databases}
Six popular image databases are used in our experiments, including Extended Yale Database B (ExYaleB)~\cite{Lee2005}, Columbia Object Image Library (COIL20)~\cite{Nene1996a}, Columbia Object Image Library (COIL100)~\cite{Nene1996b}, USPS~\cite{Hull1994}, MNIST~\cite{Lecun1998}, and Covtype~\cite{Blackard1998}. We give the details of these databases as follows:

\begin{itemize}

\item The ExYaleB database contains $2,414$ frontal face images of $38$ subjects and around 64 near frontal images under different illuminations per individual, where each image is manually cropped and normalised to the size of $32 \times 32$ pixels~\cite{Cai2011}.

\item The COIL20 and COIL100 databases contain 20 and 100 objects, respectively. The images of each object were taken $5$ degrees apart as the object is rotated on a turntable and each object has $72$ images. The size of each image is $32 \times32$ pixels, with $256$ grey levels per pixel~\cite{Cai2011}.

\item The USPS handwritten digit database\footnote{The USPS database and MNIST database used in this paper are download from
\url{http://www.cad.zju.edu.cn/home/dengcai/Data/MLData.html}.} includes ten classes ($0 - 9$ digit characters) and 11,000 samples in total. We use a popular subset contains $9,298$ handwritten digit images for the experiments, and all of these images are normalised to the size of $16 \times 16$ pixels. In the experiment, we select $200$ samples of each subject from the database randomly by following the strategy in ~\cite{Cheng2010}.

\item The MNIST handwritten digit database includes ten classes ($0 - 9$ digit characters) and $60,000$ samples in total. We use first $10,000$ handwritten digit images of the training subset to conduct the experiments, and all of these images are normalised to the size of $28 \times 28$ pixels. In the experiment, we also select $200$ samples of each subject from the database randomly to evaluate the performance of different algorithms.

\item The Covtype database is a large-scale database, which consists of
    $581,012$ samples of $7$ subjects, and each sample has $54$ attribute variables. This database is developed to investigate the performance of algorithms on predicting forest cover type from cartographic variables only (no remotely sensed data). Independent variables were derived from data originally obtained from the US Geological Survey (USGS) and USFS data. In the experiments, we normalise the attribute values to the interval $[0, 1]$.
\end{itemize}

%
%
%
%
%

\subsection{Baselines and Evaluation Metrics}
We compare KTRR\footnote{The source code is available at
\url{https://liangli-zhen.github.io/code/KTRR.zip}.}
 with the state-of-the-art subspace clustering approaches, including truncated regression representation (TRR)~\cite{Peng2016}, two versions of least squares regression (LSR1, LSR2)~\cite{Lu2013}, kernel low-rank representation (KLRR)~\cite{Xiao2015}, kernel sparse subspace clustering (KSSC)~\cite{patel2014kernel}, Latent low-rank representation (LatLRR)~\cite{Liu2011}, low-rank representation with $\ell_1$-norm (LRR1)~\cite{Liu2013}, low-rank representation with $\ell_{21}$-norm (LRR2)~\cite{Liu2013}, sparse subspace clustering (SSC)~\cite{Elhamifar2013}, sparse manifold clustering and embedding (SMCE)~\cite{Elhamifar2011}, local subspace analysis (LSA)~\cite{yan2006general}, and standard spectral clustering (SC)~\cite{Ng2002}, on four real-world databases. To evaluate the performance of the extension of KTRR (EKTRR) on large-scale set, we compare it with seven scalable clustering algorithms (SSSC~\cite{peng2013scalable}, SLRR~\cite{Peng2015b}, SLSR~\cite{Peng2015b}, KASP~\cite{Yan2009FAS}, Nystr\"{o}m~\cite{Chen2011parallel}, SEC~\cite{nie2011spectral}, and AKK~\cite{Chitta2011AKK}). Furthermore, we also compare the proposed method with currently developed deep subspace clustering methods including deep adversarial subspace clustering (DASC)~\cite{zhou2018deep}, deep subspace clustering network with $\ell_2$-norm regularization on (DSC-Net-L2)~\cite{Ji2017deep}, deep subspace clustering network with $\ell_1$-norm regularization on (DSC-Net-L1)~\cite{Ji2017deep}, structured autoencoders (StructAE)~\cite{peng2018stru}, and SSC with pre-trained convolutional autoencoder features (AE+SSC).

Four popular metrics are adopted to evaluate the subspace clustering quality, \textit{i.e.}, accuracy (AC)~\cite{Cheng2010}, normalized mutual information (NMI)~\cite{Cheng2010}, the adjusted rand index (ARI)~\cite{Hubert1985}, and F-Score~\cite{Goutte2005}. The values of these four metrics are higher if the method works better. The values of these four metrics are equal to $1$ indicates the predict result is perfectly matching with the ground truth, whereas $0$ indicates totally mismatch.

\subsection{Visualisation of Representation and Similarity Matrices}
\begin{figure*}[htpb]
\centering
\subfigure[]{ \label{coef}
\includegraphics[width=0.3\linewidth]{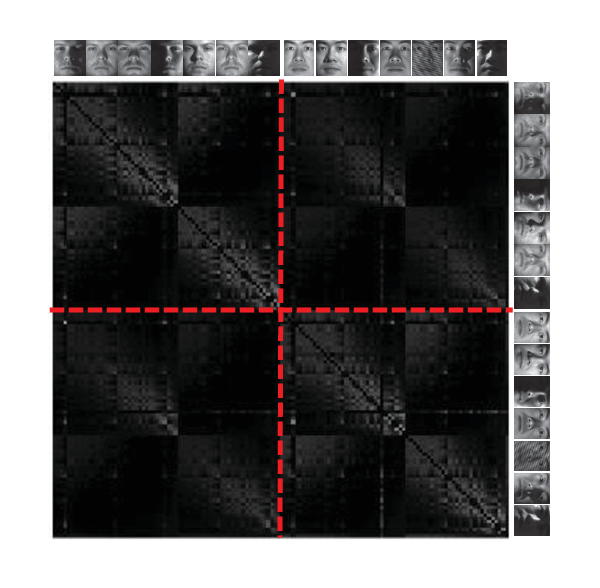}}
\subfigure[]{ \label{sym}
\includegraphics[width=0.3\linewidth]{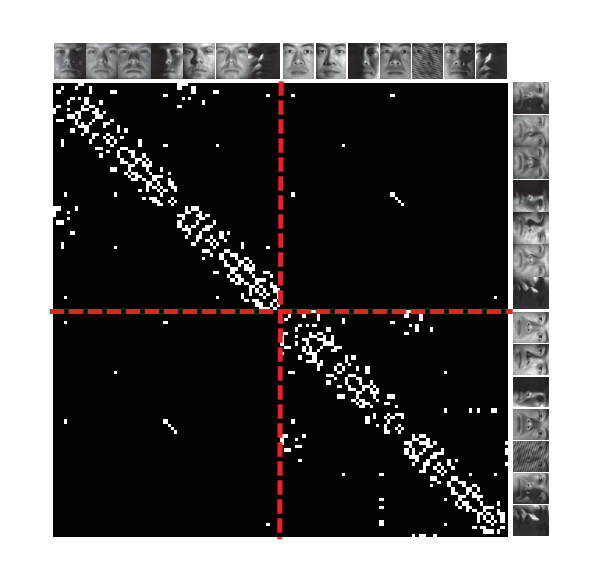}}
\caption{The visualisation of the representation matrix and the similarity matrix on $128$ facial images. They belong to the first $2$ subjects in ExYaleB database. (a) The representation matrix in (\ref{eq3.9}). (b) The similarity matrix obtained by our algorithm. The experiment was carried out on the first two subjects of ExYaleB. The top rows and the right columns illustrate some images of these two subjects. The dotted lines split each matrix into four parts. The upper-left part: the similarity relationship among the $64$ images of the first subject. The bottom-right part: the similarity relationship among the $64$ images of the second subject. The upper-right part and the bottom-left part: the similarity relationship among the images from different subjects. From the connection results, it is easy to find that most of the non-zero elements are located in the upper-left part and the bottom-right part, which means that our method reflects the relationships among the samples from different subjects very well.}
\label{fig:5.0}
\end{figure*}

Before evaluating the clustering performance of the proposed method, we demonstrate the visualisation results of the coefficient matrix of KTRR with the Gaussian kernel and the obtained similarity matrix. We conduct the experiment on the first $128$ facial images of the ExYaleB database, in which the first $64$ samples of which belong to the first subject, and the other $64$ samples belong to the second subject. We set the parameters as $\lambda = 5$ and $\eta = 4$. The representation matrix $\mathbf{C}^*$ in (\ref{eq3.9}) and the constructed similarity matrix $\mathbf{W}$ in (\ref{eq3.13}) are shown in Fig.~\ref{coef} and Fig.~\ref{sym},  respectively.

From Fig.~\ref{coef}, we can see that most of the non-zero elements are located in the upper-left part and the bottom-right part, but there still exist some non-zero elements in the upper-right part and the bottom-left part. That is to say, the connections among the same subject are much stronger than that among different subjects, while there still exist numbers of trivial connections among the samples from different subjects since the samples from these subjects are all facial images, which have some common characteristics.

It is well known that an ideal similarity matrix for the spectral clustering algorithm is a block diagonal matrix, \textit{i.e.}, the connections should only exist among the data points from the same cluster~\cite{ Liu2013, Peng2016, patel2014kernel, Xiao2015}, such that a hard thresholding operation has been executed. From the result of the similarity matrix $\mathbf{W}$ (in Fig.~\ref{sym}), we find that:
\begin{itemize}
  \item Most of the bright spots lie in the diagonal blocks of the similarity matrix, \textit{i.e.}, the strong connections exist among the samples from the same subject;
  \item Our method reveals the latent structure of data that these images belong to two subjects. There exist only a few bright spots in the upper-right part and the bottom-left part of the obtained similarity matrix, \textit{i.e.}, the trivial connections among the samples from different subjects have been mostly removed by using the thresholding processing;

  \item The obtained similarity matrix is a symmetric matrix, which can be directly used for subspace clustering under the framework of the spectral clustering~\cite{Ng2002}.
\end{itemize}

\subsection{Different Kernel Functions}

\begin{table*}[htpb]\scriptsize
\caption{Clustering performance (mean $\pm$ sd, \%) comparison of different kernel functions used in the proposed method with $10$ runs, in each run the k-means clustering step is repeated $500$ times. The parameter of $\sigma$ is setting as the mean of the distances between all the samples. The best mean results in different metrics are in bold.}
\label{tab8}
\centering
\begin{tabular}{ll||llllll}
\hline
    \multicolumn{2}{l||}{Function $\kappa(\mathbf{x}_i, \mathbf{x}_j)$} & \multicolumn{1}{l}{$(\mathbf{x}_i^T\mathbf{x}_j)^3$} & $(\mathbf{x}_i^T\mathbf{x}_j)^2$ & $e^{-\frac{\|\mathbf{x}_i - \mathbf{x}_j\|^2}{\sigma^2}}$ & $e^{-\frac{\|\mathbf{x}_i - \mathbf{x}_j\|}{\sigma}}$  & $\frac{1}{\|\mathbf{x}_i - \mathbf{x}_j\|^2}$&$\frac{1}{\|\mathbf{x}_i - \mathbf{x}_j\|}$\\
\hline
\hline
    \multirow{4}[0]{*}{ExYaleB}    &AC    &10.62 $\pm$ 0.58    &65.02 $\pm$ 3.22    &\textbf{84.82 $\pm$ 5.75}    &77.43 $\pm$ 3.70    &78.84 $\pm$ 3.90    &76.39 $\pm$ 3.76\\
    &NMI    &14.03 $\pm$ 0.87    &74.85 $\pm$ 1.59    &\textbf{89.52 $\pm$ 2.43}    &81.07 $\pm$ 1.68    &86.21 $\pm$ 0.74    &80.04 $\pm$ 1.36\\
    &ARI    &1.32 $\pm$ 0.17    &50.04 $\pm$ 5.77    &\textbf{77.09 $\pm$ 5.84}    &62.08 $\pm$ 4.18    &72.92 $\pm$ 2.41    &58.82 $\pm$ 3.88\\
    &F-Score    &4.37 $\pm$ 0.25    &51.47 $\pm$ 5.49    &\textbf{77.72 $\pm$ 5.66}    &63.15 $\pm$ 4.03    &73.66 $\pm$ 2.34    &60.01 $\pm$ 3.71\\
 \hline
    \multirow{4}[0]{*}{COIL20}    &AC    &70.99 $\pm$ 6.79    &85.03 $\pm$ 0.92    &90.25 $\pm$ 6.13    &\textbf{91.81 $\pm$ 0.00}    &90.56 $\pm$ 0.00    &\textbf{91.81 $\pm$ 0.00}\\
    &NMI    &81.04 $\pm$ 3.00    &94.11 $\pm$ 0.27    &94.71 $\pm$ 2.75    &\textbf{96.24 $\pm$ 0.00}    &95.23 $\pm$ 0.00    &96.06 $\pm$ 0.00\\
    &ARI    &61.55 $\pm$ 8.39    &82.98 $\pm$ 0.07    &88.04 $\pm$ 5.49    &\textbf{91.14 $\pm$ 0.00}    &89.11 $\pm$ 0.00    &90.89 $\pm$ 0.00\\
    &F-Score    &63.71 $\pm$ 7.71    &83.92 $\pm$ 0.07    &88.65 $\pm$ 5.19    &\textbf{91.60 $\pm$ 0.00}    &89.66 $\pm$ 0.00    &91.36 $\pm$ 0.00\\
 \hline

    \multirow{4}[0]{*}{USPS} & AC    & 80.38 $\pm$ 19.04 & 72.31 $\pm$ 18.06 & \textbf{81.36 $\pm$ 14.93} & 74.97 $\pm$ 12.40 & 79.62 $\pm$ 17.69 & 73.64 $\pm$ 14.46 \\
          & NMI   & 76.08 $\pm$ 9.75 & 67.38 $\pm$ 14.41 & \textbf{78.04 $\pm$ 6.63} & 75.59 $\pm$ 9.22 & 74.18 $\pm$ 13.08 & 74.58 $\pm$ 8.01 \\
          & ARI   & 70.10 $\pm$ 15.43 & 59.48 $\pm$ 17.11 & \textbf{71.73 $\pm$ 12.67} & 65.98 $\pm$ 13.81 & 68.32 $\pm$ 19.86 & 64.62 $\pm$ 14.44 \\
          & F-Score & 73.25 $\pm$ 13.45 & 63.78 $\pm$ 15.06 & \textbf{74.69 $\pm$ 11.13} & 69.72 $\pm$ 11.83 & 71.68 $\pm$ 17.47 & 68.53 $\pm$ 12.38 \\
 \hline
    \multirow{4}[0]{*}{MNIST} & AC    & 65.58 $\pm$ 11.65 & \textbf{66.48 $\pm$ 14.54} & 63.97 $\pm$ 8.11 & 59.21 $\pm$ 14.27 & 61.62 $\pm$ 12.43 & 62.06 $\pm$ 9.67 \\
          & NMI   & 64.27 $\pm$ 6.25 & 63.49 $\pm$ 6.50 & \textbf{66.81 $\pm$ 4.43} & 63.54 $\pm$ 12.66 & 64.00 $\pm$ 9.32 & 65.33 $\pm$ 5.59 \\
          & ARI   & 51.62 $\pm$ 10.21 & 51.54 $\pm$ 11.54 & \textbf{52.63 $\pm$ 5.04} & 48.62 $\pm$ 16.10 & 50.18 $\pm$ 11.25 & 51.20 $\pm$ 7.18 \\
          & F-Score & 56.70 $\pm$ 8.92 & 56.61 $\pm$ 10.09 & \textbf{57.92 $\pm$ 4.16} & 54.50 $\pm$ 13.94 & 55.63 $\pm$ 9.71 & 56.66 $\pm$ 6.13 \\
\hline
\end{tabular}
\end{table*}

There are many kernel functions, and some of them are commonly used, \textit{e.g.}, polynomial kernels, radial basis functions, and sigmoid kernels. To investigate the performance of the proposed method using different kernels, we study six different kernel functions. We conduct the experiments on four databases, \textit{i.e.}, ExYaleB, COIL20, USPS and MINIST. The clustering performance with different kernels are shown in Table~\ref{tab8}, from which we have the following observations:
\begin{itemize}

\item The kernel function $\kappa(\mathbf{x}_i, \mathbf{x}_j) = e^{-\frac{\|\mathbf{x}_i - \mathbf{x}_j\|^2}{\sigma^2}}$ achieves the best performance on ExYaleB and USPS, and obtains competitive performance compared with the kernel functions $\kappa(\mathbf{x}_i, \mathbf{x}_j) = (\mathbf{x}_i^T\mathbf{x}_j)^2$ and $\kappa(\mathbf{x}_i, \mathbf{x}_j) = e^{-\frac{\|\mathbf{x}_i - \mathbf{x}_j\|}{\sigma}}$ on COIL20 and MNIST, respectively.

\item The performance of the proposed method with the kernel functions $\kappa(\mathbf{x}_i, \mathbf{x}_j) = (\mathbf{x}_i^T\mathbf{x}_j)^3$ and $\kappa(\mathbf{x}_i, \mathbf{x}_j) = (\mathbf{x}_i^T\mathbf{x}_j)^2$ on ExYaleB is poor. It illustrates that the facial images cannot be project into linear spaces with the mapping function corresponding to these polynomial kernel functions.

\item The performance of the proposed method with $\kappa(\mathbf{x}_i, \mathbf{x}_j) = (\mathbf{x}_i^T\mathbf{x}_j)^2$ outperforms $\kappa(\mathbf{x}_i, \mathbf{x}_j) = (\mathbf{x}_i^T\mathbf{x}_j)^3$ on COIL20, which is different from that on USPS. The same observation can be obtained on results with $\kappa(\mathbf{x}_i, \mathbf{x}_j) = e^{-\frac{\|\mathbf{x}_i - \mathbf{x}_j\|}{\sigma}}$ and $\kappa(\mathbf{x}_i, \mathbf{x}_j) = e^{-\frac{\|\mathbf{x}_i - \mathbf{x}_j\|^2}{\sigma^2}}$. It is mainly caused by the fact that the images from USPS lie in much higher nonlinear subspaces than those from the COIL20, and the functions $\kappa(\mathbf{x}_i, \mathbf{x}_j) = (\mathbf{x}_i^T\mathbf{x}_j)^3$ and $\kappa(\mathbf{x}_i, \mathbf{x}_j) = e^{-\frac{\|\mathbf{x}_i - \mathbf{x}_j\|^2}{\sigma^2}}$ induced a much more nonlinear mapping than $\kappa(\mathbf{x}_i, \mathbf{x}_j) = (\mathbf{x}_i^T\mathbf{x}_j)^2$ and $\kappa(\mathbf{x}_i, \mathbf{x}_j) = e^{-\frac{\|\mathbf{x}_i - \mathbf{x}_j\|}{\sigma}}$, respectively.

\item The selection of different kernels results in a significant difference in the subspace clustering performance both on these four databases.

\end{itemize}

Since the proposed method with Gaussian kernel can obtain promising performance from the above experimental results, and it is also the most commonly used kernel~\cite{Shawe2004, Xiao2015}, we adopt the Gaussian kernel function to compute the kernel matrix for our method in the rest of the experiments.

\subsection{Parameter Analysis}
KTRR has two parameters, the tradeoff parameter $\lambda$ and the parameter $\eta$. The selection of the values of the parameters depends on the data distribution. A bigger $\lambda$ is suitable for highly corrupted databases, and $\eta$ corresponds to the dimensionality of the corresponding subspace for the mapped data points.

\begin{figure*}[hbpt]
\centering
\subfigure[]{ \label{pairwise}
\includegraphics[width=0.27\linewidth]{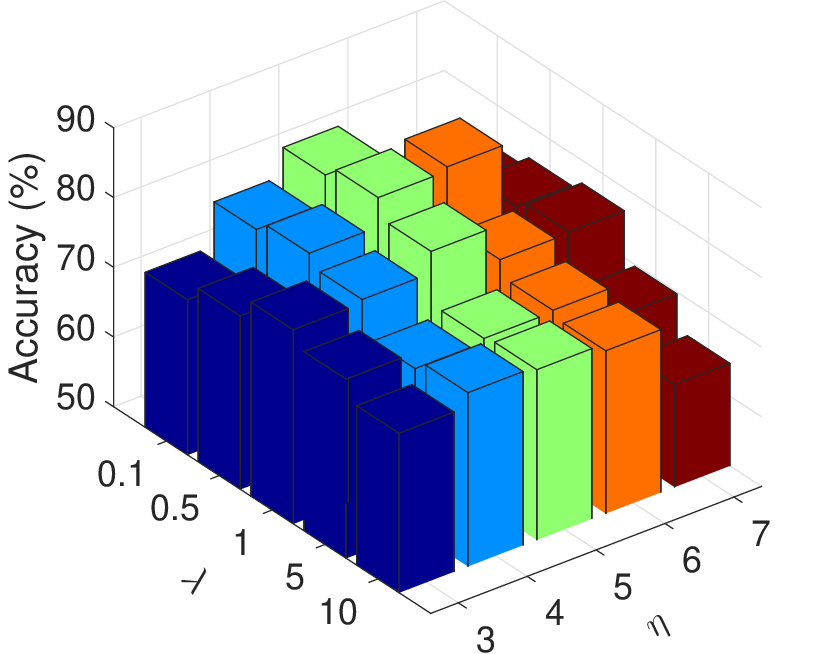}}
\subfigure[]{ \label{linear}
\includegraphics[width=0.33\linewidth]{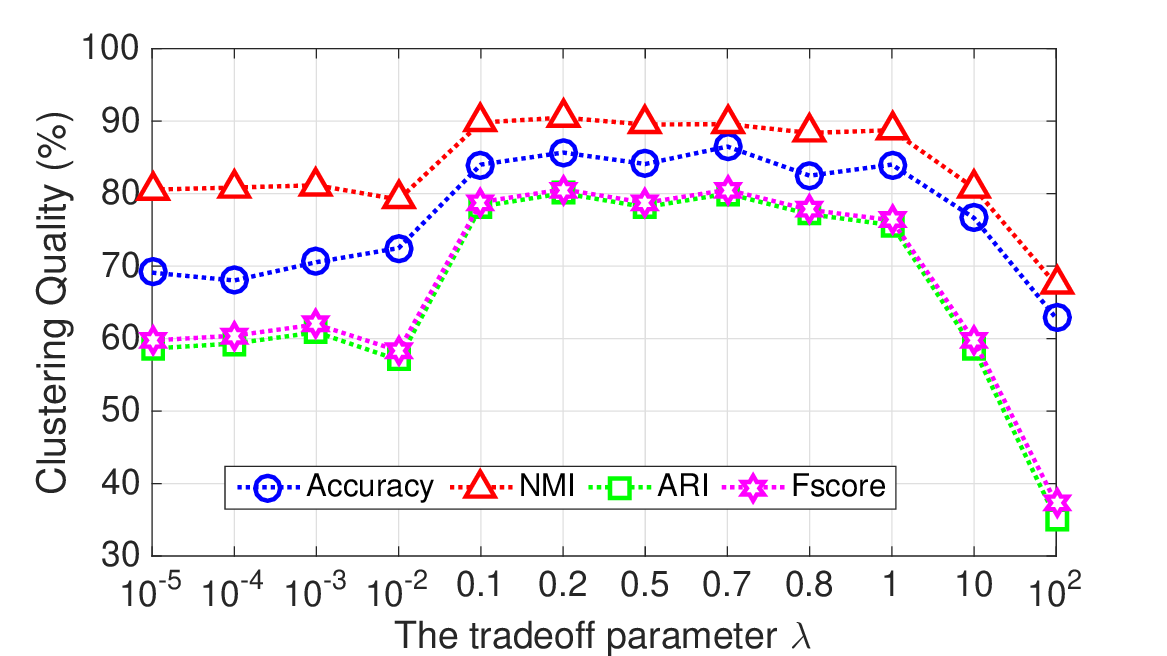}}
\subfigure[]{\label{locallylinear}
\includegraphics[width=0.33\linewidth]{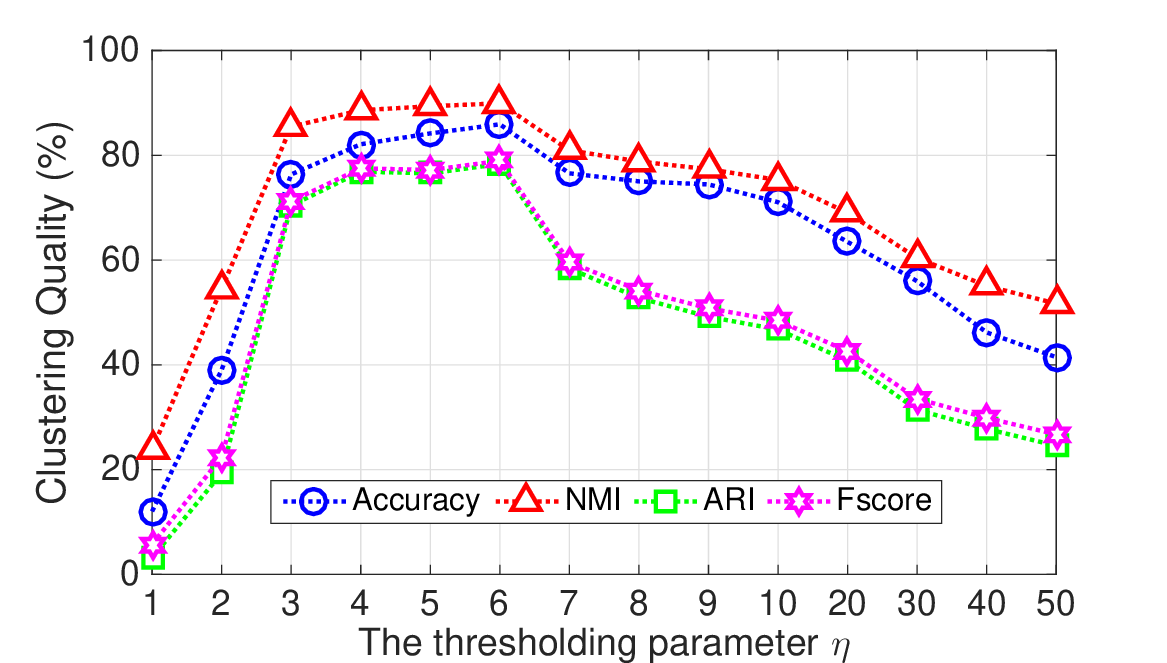}}
\caption{Clustering performance (mean of 50 runs) of the proposed method on the ExYaleB database. (a) Clustering performance of the proposed method versus different values of $\lambda$ and $\eta$. (b) Clustering performance of the proposed method versus different values of $\lambda$ under $\eta = 5$. (c) Clustering performance of the proposed method versus different values of $\eta$ under $\lambda = 0.5$.} \label{fig:5.1}
\end{figure*}

\begin{figure*}[hbpt]
\centering
\subfigure[]{
\includegraphics[width=0.27\linewidth]{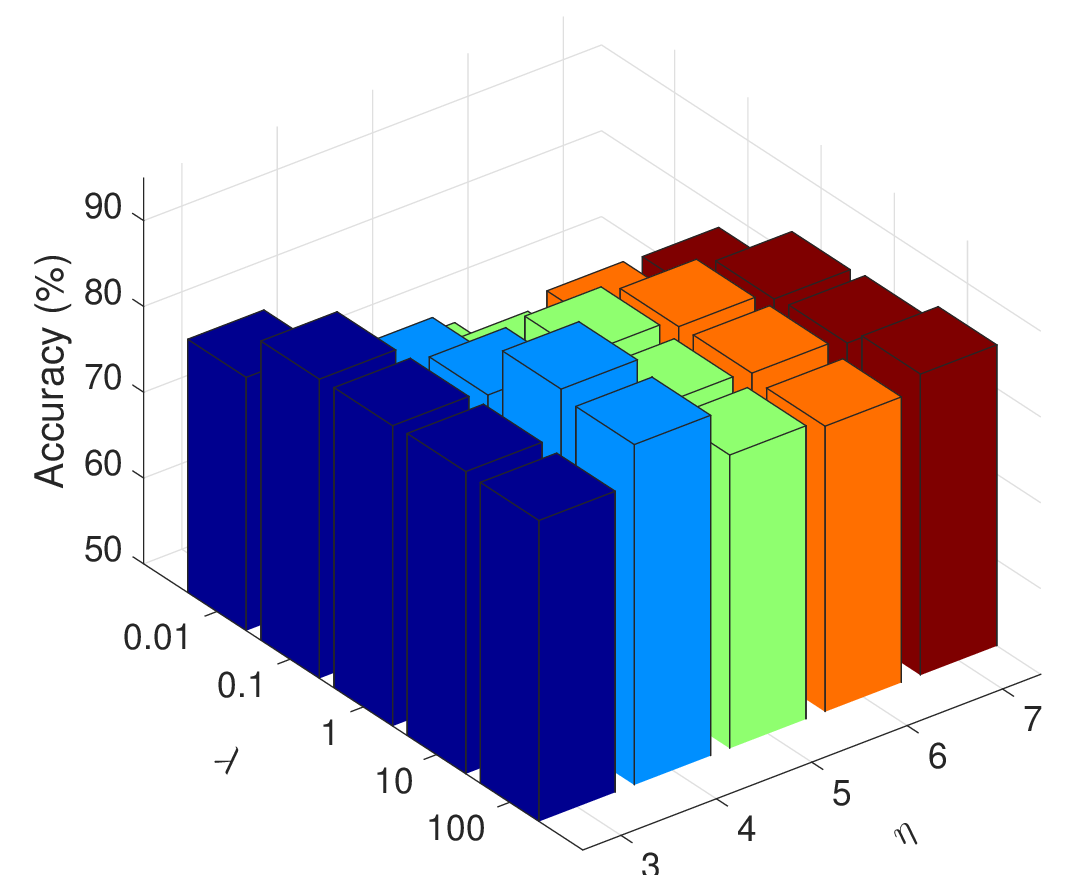}}
\subfigure[]{
\includegraphics[width=0.33\linewidth]{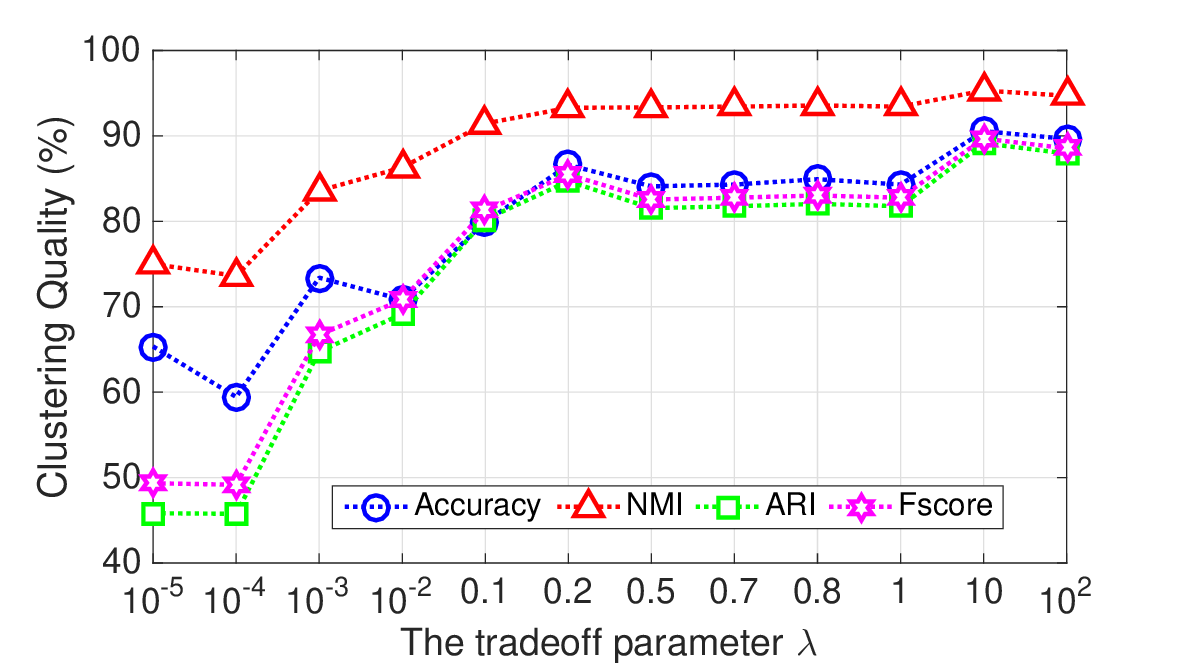}}
\subfigure[]{
\includegraphics[width=0.33\linewidth]{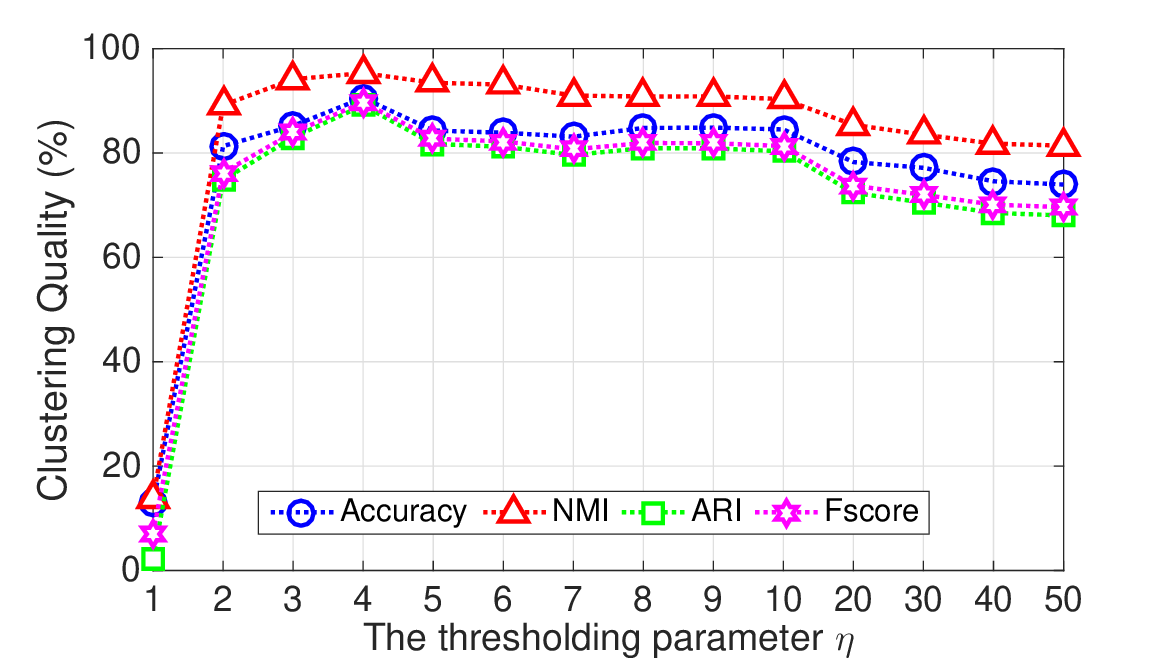}}
\caption{Clustering performance (mean of 50 runs) of the proposed method on the COIL20 database. (a) Clustering performance of the proposed method versus different values of $\lambda$ and $\eta$. (b) Clustering performance of the proposed method versus different values of $\lambda$ under $\eta = 4$. (c) Clustering performance of the proposed method versus different values of $\eta$ under $\lambda = 10$.} \label{fig:5.1b}
\end{figure*}

To evaluate the impact of $\lambda$ and $\eta$, we conduct the experiment on the ExYaleB and COIL20 databases. We set the $\lambda$ from $10^{-5}$ to $10^2$, and $\eta$ from $1$ to $50$, the results are shown in Fig.~\ref{fig:5.1} and Fig.~\ref{fig:5.1b}.

From the results, we have the following observations:
\begin{itemize}
\item KTRR achieves the best clustering performance with $\lambda$ and $\eta$ as $0.1$ and $5$ on the ExYaleB database, and $10$ and $4$ on the COIL20 database, respectively.

\item KTRR can obtain satisfactory performance with $\lambda$ from $0.1$ to $1$ on the ExYaleB database, where the values of Accuracy, NMI, ARI, and F-Score are more than $85 \%$, $90 \%$, $75 \%$, and $75 \%$, respectively, and with $\lambda$ from $0.2$ to $100$ on the COIL20 database, where the Accuracy, NMI, ARI, and F-Score are more than $80 \%$, $90 \%$, $80 \%$, and $80 \%$. The performance of KTRR is not sensitive to the parameter of $\lambda$, which enables KTRR being suitable for the real-world applications.

\item The clustering quality with $\eta$ from $3$ to $10$ on the ExYaleB and the COIL20 databases are much better than other cases. It means that the thresholding process is helpful to improve the performance of KTRR, and the dimensions of the hidden subspaces of the ExYaleB and the COIL20 databases are in the $3$ to $10$ range.
\end{itemize}

\subsection{Clustering Performance with Different Number of Subjects}
In this subsection, we investigate the clustering performance of the proposed method with a different number of subjects on the COIL100 image database. The experiments are carried out on the first $t$ classes of the database, where $t$ increases from $10$ to $100$ with an interval of $10$. The clustering results are shown in Fig.~\ref{fig:5.2}, from which we can see that:

\begin{figure}[tpb]
\centering
\includegraphics[width=0.8\linewidth]{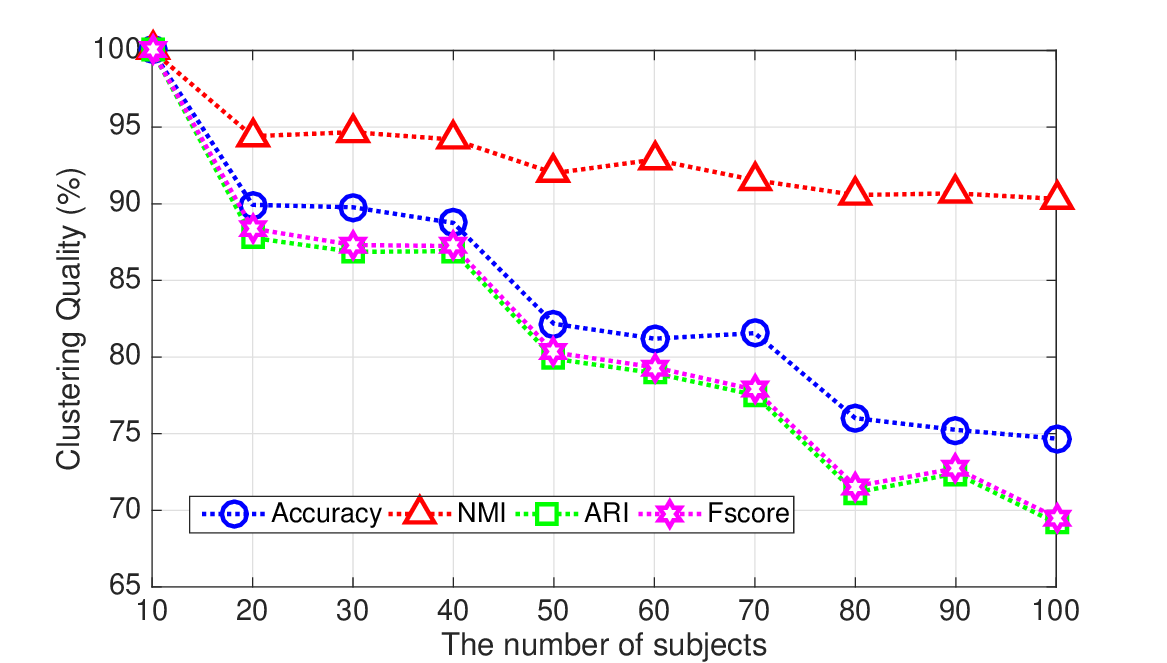}
\caption{The clustering mean quality of the proposed method on the first $t$ subjects of the COIL100 database with 50 runs.}
\label{fig:5.2}
\end{figure}

\begin{itemize}
\item In general, with the number of subjects increase, the clustering performance is decreased since the clustering difficulty is increasing with the number of subjects.
\item With an increasing number of subjects, the NMI of KTRR is changed slightly, varying from $100 \%$ to $90 \%$. The potential reason is that the NMI is robust to the data distribution (increasing subject number)~\cite{Peng2016}.
\item The proposed method obtains satisfactory performance on the COIL100 database. It achieves perfect clustering result for $t = 10$, and gets the satisfactory performance at $t = 100$ with Accuracy, NMI, ARI, and F-Score be around $74 \%$, $90 \%$, $68 \%$, and $68 \%$, respectively.
\end{itemize}

\subsection{Comparison with Existing Methods on Clean Images}
In this experiment, we compare KTRR with other $12$ state-of-the-art approaches on four different benchmark databases, \textit{i.e.}, Extended Yale Database B (ExYaleB)~\cite{Lee2005}, Columbia Object Image Library (COIL20)~\cite{Nene1996a}, USPS~\cite{Hull1994}, MNIST~\cite{Lecun1998}. The performance comparisons on COIL100 and Covtype are not provided here since several tested algorithms need a very long period of time to obtain their results on these two databases. For a fair comparison, we use the same spectral clustering framework~\cite{Ng2002} with different similarity matrices obtained by the tested methods. According to the setting in \cite{Xiao2015}, for all of the kernel-based algorithms, we adopt the commonly-used Gaussian kernel on all datasets and use the default bandwidth parameter which is set to the mean of the distances between all the samples. For each dataset, we perform each method $10$ runs, in each run the k-means clustering step is repeated $500$ times and report the mean and the standard deviation of the used metrics. The clustering results on the above four databases are shown in Table~\ref{tab3} - Table~\ref{tab6}. The best means for each database are highlighted in boldface. To have statistically sound conclusions, the Wilcoxon's rank sum test~\cite{gibbons2011nonparametric} at a $0.05$ significance level is adopted to test the significance of the differences between the results obtained by the proposed method and all other algorithms. From the results, we can obtain the following conclusions.

\begin{table}[htbp]\scriptsize
\centering
\caption{Clustering performance (mean $\pm$ sd, \%) comparisons of different methods on the ExYaleB database. The best mean results in different metrics are in bold. The ``$\dagger$'' indicates that the value of the proposed method is significantly different from all other methods at a $0.05$ level by the Wilcoxon's rank sum test.}
\begin{tabular}{l||llll}
\hline
Methods & AC          & NMI         & ARI         & F-Score      \\
\hline
\hline
KTRR      &\textbf{84.82$\pm$5.75}$\dagger$ & \textbf{89.52$\pm$2.43}$\dagger$ & \textbf{77.09$\pm$5.84}$\dagger$ & \textbf{77.72$\pm$5.66}$\dagger$ \\
TRR       & 67.04$\pm$2.93 & 72.20$\pm$2.61 & 41.07$\pm$6.91 & 43.01$\pm$6.52 \\
LSR1    & 55.56$\pm$3.36 & 58.59$\pm$1.45 & 33.24$\pm$2.15 & 35.20$\pm$2.02\\
LSR2     & 51.07$\pm$4.45 & 54.03$\pm$2.59 & 25.42$\pm$2.41 & 27.78$\pm$2.24\\
KLRR      & 52.30$\pm$4.31 & 61.98$\pm$2.45 & 36.06$\pm$3.49 & 37.87$\pm$3.37 \\
KSSC      & 58.41$\pm$3.19 & 64.41$\pm$1.10 & 32.40$\pm$5.82 & 34.59$\pm$5.38 \\
LatLRR    & 51.40$\pm$3.36 & 54.41$\pm$1.76 & 27.10$\pm$2.26 & 29.33$\pm$2.07 \\
LRR1  & 50.32$\pm$2.68 & 53.31$\pm$1.42 & 26.42$\pm$2.17 & 28.66$\pm$2.00 \\
LRR2 & 49.80$\pm$4.72 & 53.26$\pm$2.22 & 25.63$\pm$2.70 & 27.93$\pm$2.52 \\
SSC       & 52.87$\pm$5.46 & 58.02$\pm$3.44 & 24.20$\pm$4.74 & 26.83$\pm$4.34 \\
SMCE      & 48.91$\pm$3.71 & 60.22$\pm$1.28 & 30.46$\pm$3.06 & 32.54$\pm$2.84 \\
LSA       & 33.97$\pm$3.95 & 47.38$\pm$1.87 & 20.98$\pm$1.45 & 23.20$\pm$1.36 \\
SC        & 19.69$\pm$1.70 & 32.96$\pm$1.54 & 10.16$\pm$1.01 & 12.56$\pm$0.98 \\
\hline
\end{tabular}
\label{tab3}%
\end{table}

(1) Evaluation on the ExYaleB facial database
\begin{itemize}
\item The KTRR algorithm achieves the best results in the tests and gains a significant improvement over TRR. The means of Accuracy, NMI, ARI, and F-Score of KTRR are about $17 \%$, $17 \%$, $26 \%$ and $24 \%$ higher than that of TRR, $32\%$, $28\%$, $41\%$, and $40\%$ higher than that of KLRR.

\item TRR~\cite{Peng2016} outperforms LSR1~\cite{Lu2013} and LSR2~\cite{Lu2013} with a considerable gap. It means that the hard thresholding operator over the coefficient vectors has a significant impact on the performance of TRR since the main difference between them is that the former one has the hard thresholding step.

\item All representation-based methods, \textit{i.e.}, KTRR, TRR~\cite{Peng2016}, KLRR~\cite{Xiao2015}, KSSC~\cite{patel2014kernel}, LRR~\cite{Liu2013} and SSC~\cite{Elhamifar2013}, outperform the standard spectral clustering method~\cite{Ng2002}. SC is failed due to some parts of the images from different subjects are similar in ExYaleB.

\item All the linear representation methods, \textit{i.e.}, TRR~\cite{Peng2016}, LRR~\cite{Liu2013} and SSC~\cite{Elhamifar2013}, are inferior to their kernel-based extensions, \textit{i.e.}, KTRR, KLRR~\cite{Xiao2015}, and KSSC~\cite{patel2014kernel}. It means that the nonlinear representation methods are more suitable to model the ExYaleB facial images.
\end{itemize}

\begin{table}[htbp]\scriptsize
  \centering
  \caption{Clustering performance (mean $\pm$ sd, \%) comparisons of different methods on the COIL20 database. The best mean results in different metrics are in bold. The ``$\dagger$'' indicates that the value of the proposed method is significantly different from all other methods at a $0.05$ level by the Wilcoxon's rank sum test.}
 \begin{tabular}{l||llll}
\hline
Methods & AC          & NMI         & ARI         & F-Score      \\
\hline
\hline
KTRR      &\textbf{90.25$\pm$6.13}$\dagger$ & \textbf{94.71$\pm$2.75}$\dagger$ & \textbf{88.04$\pm$5.49}$\dagger$ & \textbf{88.65$\pm$5.19}$\dagger$ \\
TRR    & 84.12 $\pm$ 3.35  & 91.79 $\pm$ 0.94 & 80.72 $\pm$ 3.05  & 81.76 $\pm$ 2.80  \\
LSR1   & 67.93$\pm$0.26 & 76.98$\pm$0.40 & 59.95$\pm$0.41 & 61.96$\pm$0.39\\
LSR2   & 67.24$\pm$0.30 & 75.75$\pm$0.23 & 58.53$\pm$0.17 & 60.62$ \pm$ 0.16\\
KLRR   & 68.66 $\pm$ 5.51  & 77.93 $\pm$ 3.24 & 61.96 $\pm$ 6.98  & 63.92 $\pm$ 6.55  \\
KSSC   & 79.39 $\pm$ 8.15  & 89.50 $\pm$ 2.73 & 76.54 $\pm$ 7.13  & 77.81 $\pm$ 6.65  \\
LatLRR & 67.97 $\pm$ 3.47  & 76.78 $\pm$ 1.32 & 60.03 $\pm$ 2.42  & 62.03 $\pm$ 2.30  \\
LRR1   & 67.94 $\pm$ 7.97  & 76.45 $\pm$ 2.20 & 60.03 $\pm$ 4.94  & 62.09 $\pm$ 4.66  \\
LRR2   & 66.59 $\pm$ 3.35  & 75.33 $\pm$ 2.50 & 58.45 $\pm$ 4.21  & 60.57 $\pm$ 3.98  \\
SSC    & 69.39 $\pm$ 5.93  & 80.61 $\pm$ 2.44 & 62.14 $\pm$ 5.18  & 64.17 $\pm$ 4.80  \\
SMCE   & 76.51 $\pm$ 15.98 & 90.51 $\pm$ 5.69 & 75.20 $\pm$ 15.45 & 76.61 $\pm$ 14.41 \\
LSA    & 72.86 $\pm$ 6.67  & 81.49 $\pm$ 3.69 & 68.13 $\pm$ 5.92  & 69.74 $\pm$ 5.57  \\
SC     & 69.17 $\pm$ 3.81  & 79.43 $\pm$ 2.18 & 63.77 $\pm$ 3.88  & 65.60 $\pm$ 3.68  \\
\hline
\end{tabular}%
\label{tab4}%
\end{table}%

(2) Evaluation on the COIL20 database
\begin{itemize}

\item The KTRR algorithm obtains the Accuracy of $90.25 \%$, which is better than all other tested methods. Specifically, the Accuracy of KTRR is about $6.13\%$ higher than that of the second best method TRR, and $10.96 \%$ higher than that of the third best method KSSC.

\item All the linear representation methods, \textit{i.e.}, TRR~\cite{Peng2016}, LRR~\cite{Liu2013}, and SSC~\cite{Elhamifar2013}, are still inferior to their kernel-based extensions, \textit{i.e.}, KTRR, KLRR~\cite{Xiao2015}, and KSSC~\cite{patel2014kernel}. Their non-linear versions obtain the Accuracy improvements of $6.13 \%$, $0.68\%$, and $10 \%$, respectively. It means the images in COIL20 still lie in multiple nonlinear subspaces.

\item KLRR, LatLRR and two types of LRR methods are all inferior to the standard spectral method. It means that the mapped data of COIL20 cannot be represented by other mapped data with the low-rank constraint.
\end{itemize}

\begin{table}[htbp]\scriptsize
  \centering
\caption{Clustering performance (mean $\pm$ sd, \%) comparisons of different methods on the USPS handwriting database. The best mean results in different metrics are in bold. The ``$\dagger$'' indicates that the value of the proposed method is significantly different from all other methods at a $0.05$ level by the Wilcoxon's rank sum test.}
 \begin{tabular}{l||llll}
\hline
Methods & AC          & NMI         & ARI         & F-Score      \\
\hline
\hline
KTRR   & \textbf{81.36 $\pm$ 14.93}$\dagger$ & \textbf{78.04 $\pm$ 6.63}$\dagger$  & \textbf{71.73 $\pm$ 12.67}$\dagger$ & \textbf{74.69 $\pm$ 11.13}$\dagger$ \\
TRR   & 67.98$\pm$14.18 & 71.72$\pm$6.59 & 59.76$\pm$12.36 & 64.17$\pm$10.61 \\
LSR1 & 61.80$\pm$8.59 & 60.22$\pm$7.00 & 47.42$\pm$11.78 & 52.97$\pm$10.43\\
LSR2 & 62.54$\pm$9.76 & 61.37$\pm$4.38 & 48.57$\pm$8.29 & 53.99$\pm$7.28\\
KLRR   & 70.72 $\pm$ 2.63  & 66.23 $\pm$ 3.35  & 57.21 $\pm$ 3.76  & 61.64 $\pm$ 3.41  \\
KSSC   & 75.17 $\pm$ 2.89  & 73.97 $\pm$ 2.41  & 65.11 $\pm$ 3.67  & 68.76 $\pm$ 3.28  \\
LatLRR & 70.97 $\pm$ 4.10  & 66.55 $\pm$ 4.37  & 57.20 $\pm$ 4.51  & 61.59 $\pm$ 4.11  \\
LRR1   & 70.16 $\pm$ 4.29  & 66.69 $\pm$ 4.63  & 57.18 $\pm$ 4.79  & 61.58 $\pm$ 4.34  \\
LRR2   & 70.91 $\pm$ 3.96  & 66.87 $\pm$ 4.35  & 57.50 $\pm$ 4.48  & 61.85 $\pm$ 4.08  \\
SSC    & 26.86 $\pm$ 13.39 & 20.93 $\pm$ 13.44 & 9.95 $\pm$ 13.56  & 24.07 $\pm$ 7.62  \\
SMCE   & 73.77 $\pm$ 7.85  & 71.29 $\pm$ 9.69  & 62.08 $\pm$ 13.10 & 66.10 $\pm$ 11.63 \\
LSA    & 68.51 $\pm$ 8.95  & 64.80 $\pm$ 7.93  & 55.58 $\pm$ 9.00  & 60.30 $\pm$ 7.99  \\
SC     & 70.79 $\pm$ 8.58  & 62.72 $\pm$ 4.55  & 53.55 $\pm$ 5.24  & 58.25 $\pm$ 4.67\\
\hline
\end{tabular}%
\label{tab5}%
\end{table}%

(3) Evaluation on the USPS handwriting database
\begin{itemize}

\item The KTRR algorithm achieves the best results in the tests. The value of Accuracy of KTRR is about $21\%$ higher than that of the TRR, $10 \%$ higher than that of KLRR, and $6 \%$ higher than that of KSSC. The performance results of KTRR on NMI, ARI, and F-Score are also higher than other methods.

\item All the linear representation methods, \textit{i.e.}, TRR~\cite{Peng2016}, LRR~\cite{Liu2013}, and SSC~\cite{Elhamifar2013}, are inferior to their kernel-based extensions, \textit{i.e.}, KTRR, KLRR~\cite{Xiao2015}, and KSSC~\cite{patel2014kernel}. The performance improvement is considerable, \textit{e.g.}, the Accuracy value of KSSC is about $44 \%$ higher than that of SSC.

\item SSC is inferior to LRR, while its kernel-based extension KSSC achieves a good performance. The implicit transformation on the USPS images makes the mapped data points to be much better represented with each other in a sparse representation form. The performance of SSC is poor on this database. It is mainly because the images in each group are not sufficient, which makes SSC result in a wrong representation for the data and suffer a low clustering accuracy.

\end{itemize}

\begin{table}[htbp]\scriptsize
  \centering
\caption{Clustering performance (mean $\pm$ sd, \%) comparisons of different methods on the MNIST handwriting database. The best mean results in different metrics are in bold. The ``$\dagger$'' indicates that the value of the proposed method is significantly different from all other methods at a $0.05$ level by the Wilcoxon's rank sum test.}
 \begin{tabular}{l||llll}
\hline
Methods & AC          & NMI         & ARI         & F-Score      \\
\hline
\hline
KTRR   & \textbf{63.97 $\pm$ 8.11}$\dagger$  & \textbf{66.81 $\pm$ 4.43}$\dagger$  & \textbf{52.63 $\pm$ 5.04}$\dagger$  & \textbf{57.92 $\pm$ 4.16}$\dagger$  \\
TRR   & 54.05$\pm$6.40 & 54.79$\pm$5.04 & 39.72$\pm$6.66 & 46.38$\pm$5.75\\
LSR1   & 50.87$\pm$4.90 & 45.70$\pm$4.09 & 32.92$\pm$3.44 & 39.78$\pm$3.12\\
LSR2    & 46.77$\pm$5.15 & 43.37$\pm$4.20 & 29.59$\pm$4.12 & 36.90$\pm$3.52\\
KLRR   & 61.31 $\pm$ 6.14  & 60.07 $\pm$ 5.86  & 47.46 $\pm$ 5.77  & 52.99 $\pm$ 4.92  \\
KSSC   & 57.13 $\pm$ 15.57 & 59.43 $\pm$ 13.06 & 45.02 $\pm$ 16.11 & 50.99 $\pm$ 14.05 \\
LatLRR & 14.48 $\pm$ 8.37  & 4.04 $\pm$ 7.96   & 1.26 $\pm$ 4.47   & 18.13 $\pm$ 2.74  \\
LRR1   & 18.08 $\pm$ 10.57 & 6.76 $\pm$ 11.18  & 2.80 $\pm$ 6.50   & 18.42 $\pm$ 4.13  \\
LRR2   & 18.52 $\pm$ 12.34 & 7.53 $\pm$ 14.97  & 3.27 $\pm$ 8.64   & 18.43 $\pm$ 5.15  \\
SSC    & 22.02 $\pm$ 20.53 & 14.58 $\pm$ 24.25 & 6.16 $\pm$ 16.06  & 21.67 $\pm$ 9.34  \\
SMCE   & 61.66 $\pm$ 5.59  & 59.08 $\pm$ 3.99  & 46.81 $\pm$ 5.80  & 52.39 $\pm$ 5.05  \\
LSA    & 63.03 $\pm$ 7.46  & 61.94 $\pm$ 5.78  & 49.50 $\pm$ 8.19  & 54.78 $\pm$ 7.28  \\
SC     & 55.11 $\pm$ 5.45  & 48.80 $\pm$ 5.30  & 36.95 $\pm$ 5.62  & 43.38 $\pm$ 5.01  \\
\hline
\end{tabular}%
\label{tab6}%
\end{table}%

(4) Evaluation on the MNIST handwriting database
\begin{itemize}

\item The proposed method KTRR achieves the best clustering result and obtains a significant improvement of $31.78 \%$ at Accuracy on TRR. The indexes NMI, ARI, and F-Score of KTRR are also higher than all other tested methods.

\item All the linear representation methods, \textit{i.e.}, TRR~\cite{Peng2016}, LRR~\cite{Liu2013}, and SSC~\cite{Elhamifar2013}, are inferior to their kernel-based extensions, \textit{i.e.}, KTRR, KLRR~\cite{Xiao2015}, and KSSC~\cite{patel2014kernel}. Especially, LRR results in poor performance on this database, while its kernel-based version, KLRR, obtains much better clustering quality regarding Accuracy, NMI, ARI, and F-Score. It demonstrates the advantage of the kernel-based methods when dealing with high-dimensional data.

\item KTRR, KLRR, SMCE and LSA achieve the best clustering results on the MNIST handwriting images compared with other methods. However, the performances of all the test methods are not well. A more suitable kernel function may help these methods to get better performance on this type of databases.

\end{itemize}

\subsection{Comparison with Existing Methods on Corrupted Images}

\begin{figure*}[bpt]
\centering
\subfigure[]{ \label{noisegausianac}
\includegraphics[width=0.45\linewidth]{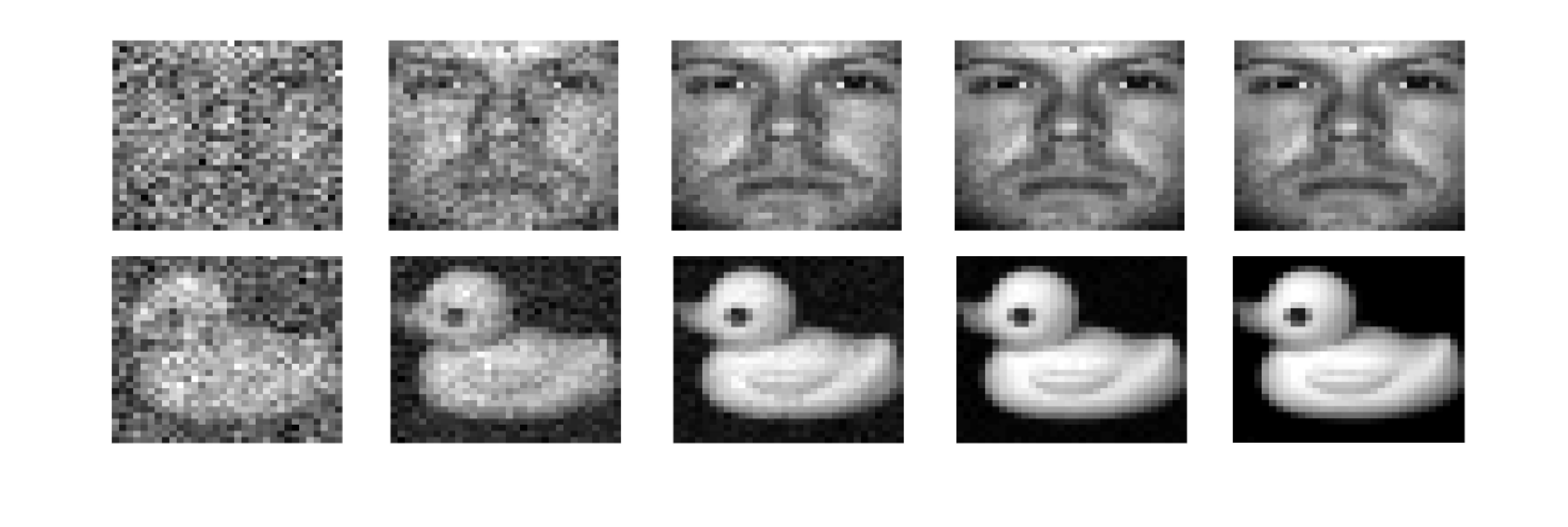}}
\subfigure[]{ \label{noisegausiannmi}
\includegraphics[width=0.45\linewidth]{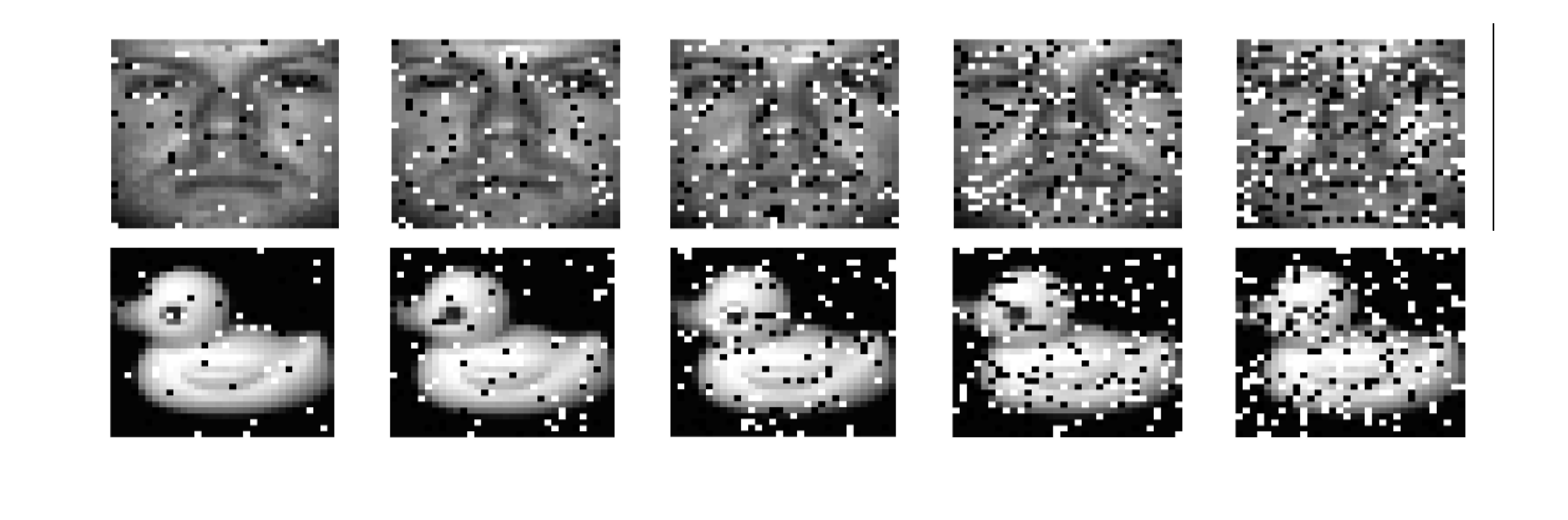}}
\caption{(a) The corrupted samples with additive Gaussian noises under SNR equals $10, 20, 30, 40, 50$ $dB$ from left to right. (b) The corrupted samples with pepper and salt noises under the ratio of affected pixels equals $5 \%, 10 \%, 15 \%, 20 \%, 25 \%$ from left to right.}
\label{fig:5.1A}
\end{figure*}

\begin{figure*}[bpt]
\centering
\subfigure[]{ \label{noisegausianac}
\includegraphics[width=0.45\linewidth]{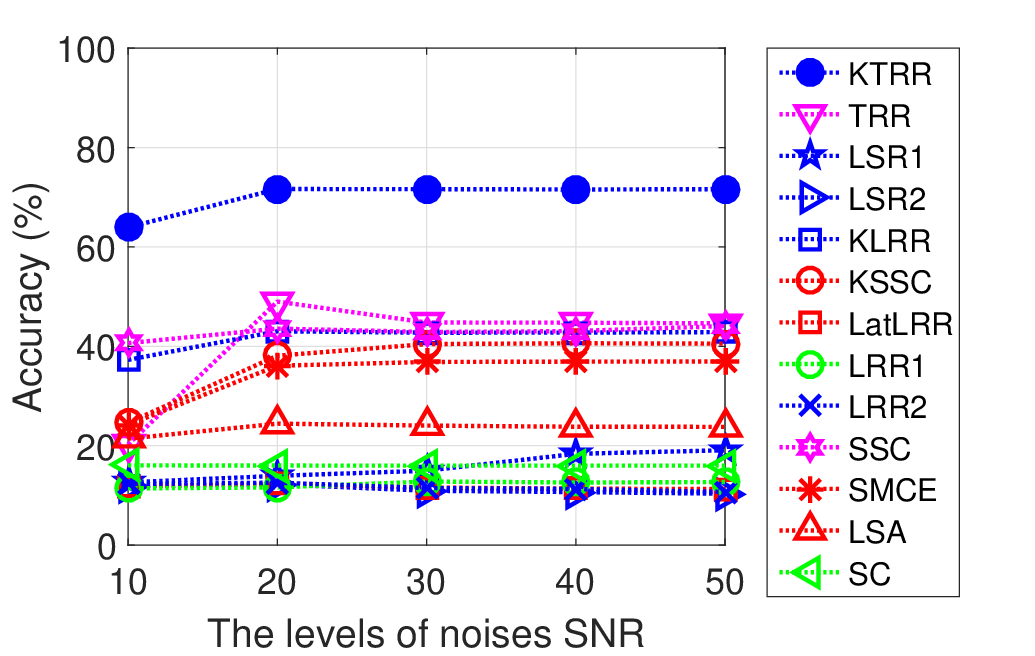}}
\subfigure[]{ \label{noisegausiannmi}
\includegraphics[width=0.45\linewidth]{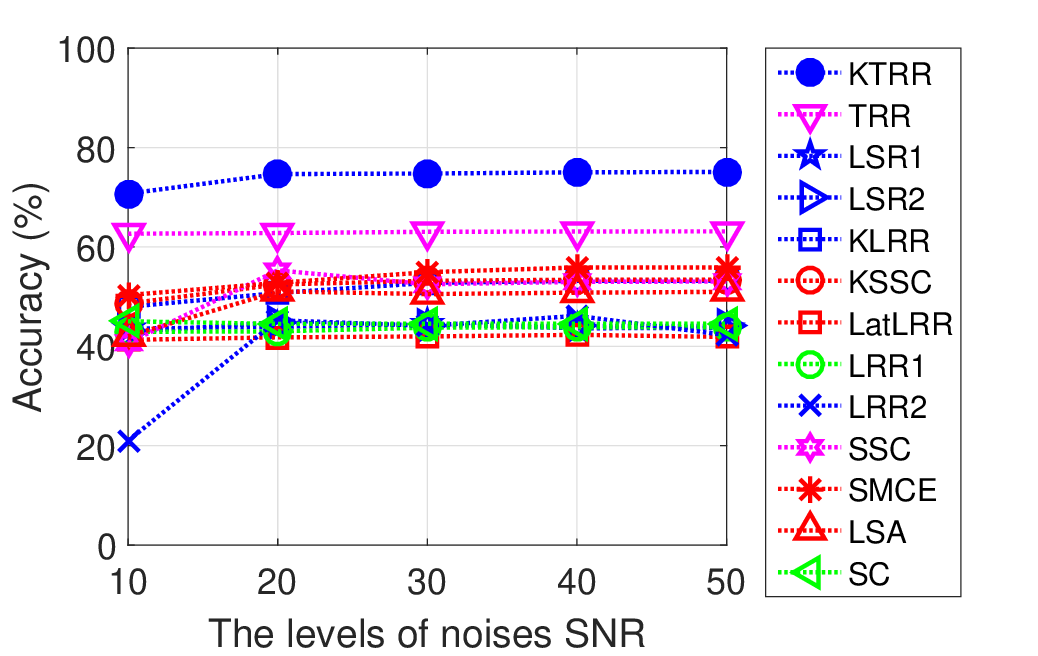}}
\caption{The clustering accuracy (mean of 50 runs) on images with different levels of additive Gaussian noises. (a) The clustering accuracy on the ExYaleB database. (b) The clustering accuracy on the COIL20 database. }
 \label{fig:5.01}
\end{figure*}

\begin{figure*}[bpt]
\centering
\subfigure[]{
\label{noisegausianac}
\includegraphics[width=0.45\linewidth]{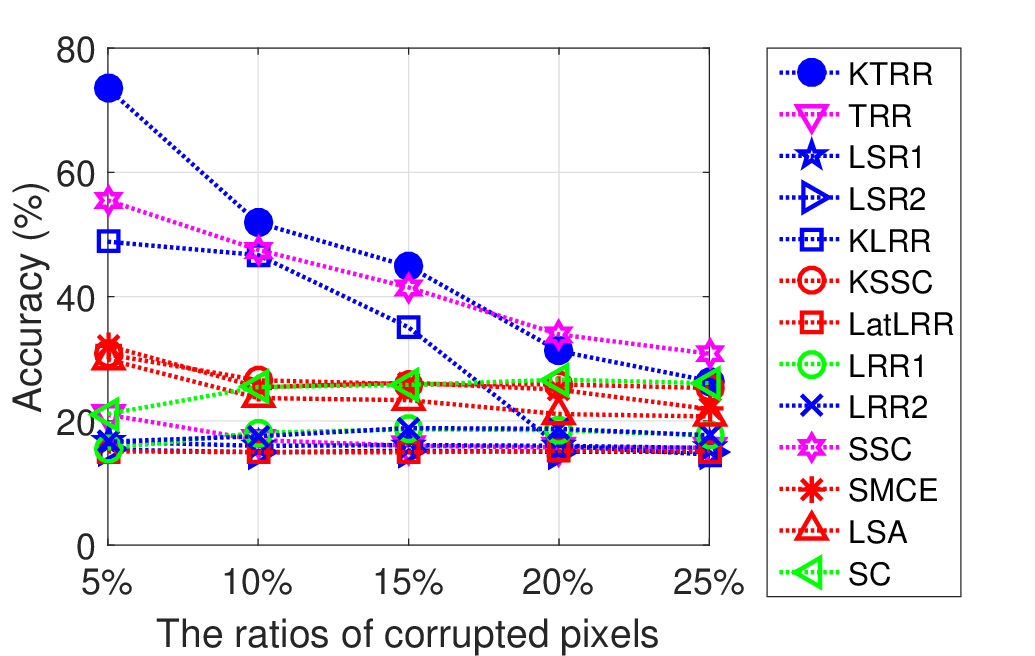}}
\subfigure[]{ \label{noisegausiannmi}
\includegraphics[width=0.45\linewidth]{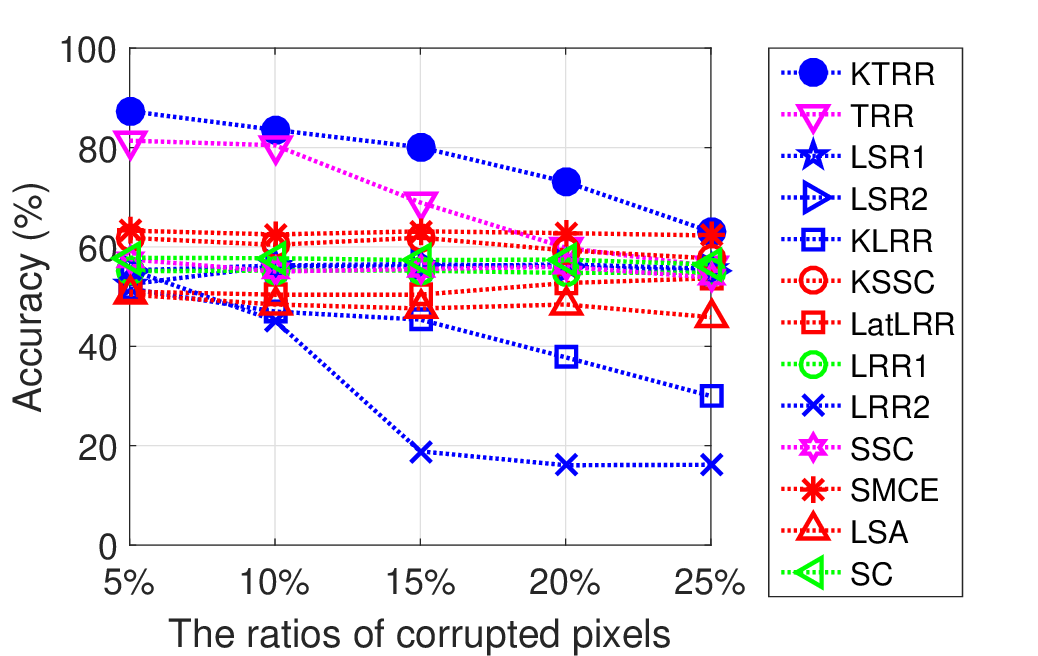}}
\caption{The clustering accuracy (mean of 50 runs) on images with different ratios of pepper \& salt corruptions. (a) The clustering accuracy on the ExYaleB database. (b) The clustering accuracy on the COIL20 database.}
 \label{fig:5.02}
\end{figure*}
To evaluate the robustness of the proposed method, we conduct the experiments on the first $10$ subjects of the COIL20 database and ExYaleB database respectively. All used images are corrupted by additive white Gaussian noises or random pixel corruptions. Some corrupted image samples under different levels of noises are as shown in Fig.~\ref{fig:5.1A}. For additive Gaussian noises, we add the noises with SNR $=10, 20, 30, 40, 50 dB$; For the random pixel corruptions, we adopt the pepper \& salt noises with the ratios of affected pixels being $5 \%, 10 \%, 15 \%, 20 \%, 25 \%$.

The clustering quality of the compared methods on the two databases with Gaussian noises is shown in Fig.~\ref{fig:5.01}, from which we can get the following observations:
\begin{itemize}
\item The proposed KTRR is considerably more robust than other methods under Gaussian noises. Specifically, KTRR obtains the Accuracy around $80 \%$ under SNR equals $10 dB$ on COIL20, which is much higher than all other tested algorithms, especially SC, LSA, LRR1, LRR2, LSR1, and LSR2.

\item Most of these spectral-based methods are relatively robust against Gaussian noises. While the performance of LRR1, LRR2, and LatLRR are sharply deteriorated on these two databases. The main reason may be that the additional Gaussian noises have destroyed the underlie the low-rank structure of the representation matrix.

\item The accuracy of all tested methods on the COIL20 database is higher than that on the ExYaleB database. It is consistent with the result of that on clean images.

\end{itemize}

The clustering quality of the compared methods on the images with randomly corruptions is shown in Fig.~\ref{fig:5.02}, from which we obtain that:
\begin{itemize}

\item The KTRR algorithm is robust to the random pixel corruptions on COIL20. It achieves the best results under the ratio of affected pixels equals $5 \%$ to $15 \%$. It obtains the Accuracy around $60 \%$ under the ratio of affected pixels equals $25 \%$ on the COIL20 database, which is a very challenging situation that we can see in Fig.~\ref{fig:5.1A}. However, the Accuracy of KTRR drops severely with the increase of the ratios of corrupted pixels, and lower than that of SSC under $20 \%$ and $25 \%$ of corrupted pixels on ExYaleB. The performance of SSC does not drops sharply, because SSC adopts a $\ell_1$-norm constraint to handle the salt $\&$ pepper corruptions.

\item All the investigated methods perform not as well as the case with white Gaussian noises. The result is consistent with a widely-accepted conclusion that non-additive corruptions are more challenging than additive ones in pattern recognition. The pixel values of the images are changed greater under the salt $\&$ pepper corruptions than that under the Gaussian noises.

\item All of the algorithms perform much better on the COIL20 database than on the ExYaleB database. The values of Accuracy of all algorithms are lower than $40 \%$ on the ExYaleB database under $20 \%$ and $25 \%$ of corrupted pixels. From Fig.~\ref{fig:5.1A}, we find that most pixel values of the images from the COIL20 database are close to $0$ or $255$. This leads to some of the corruptions to be useless and weakens the impact to the final clustering results.
\end{itemize}

\subsection{Comparison of Computational Time Cost}

\begin{table*}[htpb]\scriptsize
\caption{Computational time (seconds) comparison of different methods on the ExYaleB, COIL20, USPS, and MNIST databases. The $t_1$ and $t_2$ denote the time cost on the similarity graph construction process and the time cost on the whole clustering process of each method respectively. The best mean results in different metrics are in bold.}
\label{tab:comp}
\centering
\begin{tabular}{l||llllllll}
\hline
\multirow{2}{*}{Dataset} & \multicolumn{2}{l}{ExYaleB}    & \multicolumn{2}{l}{COIL20}    & \multicolumn{2}{l}{USPS}       & \multicolumn{2}{l}{MNIST}      \\
                           & t1            & t2             & t1            & t2            & t1            & t2             & t1            & t2             \\
\hline
\hline
KTRR                       & 22.96         & 47.62          & 6.50          & 11.66         & 16.92         & 27.82          & 22.56         & 33.96          \\
TRR                        & 23.71         & 48.8           & 6.54          & 12.75         & 11.95         & 22.74          & 22.43         & 32.35          \\
LSR1                       & 0.44          & 91.24          & 0.19          & 14.53         & 0.32          & 16.29          & 0.35          & 17.54          \\
LSR2                       & 0.43          & 88.42          & 0.21          & 14.39         & 0.16          & 14.80          & 0.24          & 19.09          \\
KLRR                       & 45.82         & 71.26          & 16.11         & 25.55         & 29.41         & 39.93          & 34.20         & 44.64          \\
KSSC                       & 5512.68       & 5543.4         & 1466.12       & 1472.07       & 2752.97       & 2763.19        & 5742.89       & 5753.42        \\
LatLRR                     & 772.44        & 806.43         & 579.01        & 584.46        & 50.05         & 58.98          & 246.35        & 270.19         \\
LRR1                       & 248.94        & 286.91         & 430.23        & 436.59        & 43.34         & 51.88          & 155.70        & 167.25         \\
LRR2                       & 270.65        & 311.34         & 454.87        & 460.07        & 49.10         & 58.90          & 172.32        & 186.82         \\
SSC                        & 2301.75       & 2313.31        & 121.76        & 126.88        & 62.25         & 98.79          & 112.13        & 153.23         \\
SMCE                       & 10.15         & \textbf{45.18} & 5.76          & 10.12         & 67.44         & 76.50          & 16.78         & 25.52          \\
LSA                        & 198.48        & 229.26         & 61.14         & 66.01         & 108.67        & 120.46         & 142.71        & 154.64         \\
SC                         & \textbf{0.33} & 124.45         & \textbf{0.15} & \textbf{7.61} & \textbf{0.14} & \textbf{11.73} & \textbf{1.09} & \textbf{14.65}\\
\hline
\end{tabular}
\end{table*}
To investigate the efficiency of KTRR, we compare its computational time with that of other $12$ approaches on the clean images of four databases. The time costs for building a similarity graph ($t_1$) and the whole time cost for clustering ($t_2$) are recorded to evaluate the efficiency of compared methods.

Table~\ref{tab:comp} shows the time costs of different methods with the parameters which achieve their best results. We can see that:
\begin{itemize}

\item KTRR and TRR~\cite{Peng2016} are much faster than KSSC, SSC, KLRR, and LRR. The results are consistent with the fact that the theocratical computation complexities of KTRR and TRR are much lower than those of KSSC, SSC, KLRR, and LRR methods. The KTRR and TRR algorithms both have analytical solutions, and only one pseudo-inverse operation is required for solving the representation problems of all data points for KTRR and TRR algorithms.

\item The standard SC~\cite{Ng2002} is the fastest since its similarity graph is computed via the pairwise kernel distances among the input samples, and LSR1 and LSR2 also have a low time cost for the similarity graph construction. KSSC~\cite{patel2014kernel} is the most time-consuming method. This implies that KSSC cannot be used to handle large-scale databases directly.

\item The time cost of the proposed method is very close to that of its linear version TRR. Specifically, TRR is faster than KTRR on the ExYaleB and the USPS databases, while it is slower than KTRR on the MNIST database. They have similar time costs on the COIL20 database. The Wilcoxon's rank sum test~\cite{gibbons2011nonparametric} at a $0.05$ significance level shows there is no significant difference between the time costs of KTRR and TRR on the similarity graph construction and the whole clustering process on the tested four databases.

\end{itemize}

\subsection{Comparison between KTRR and Deep Subspace Clustering Methods}
\begin{figure}[bpt]
\centering
\includegraphics[width= 0.8\linewidth]{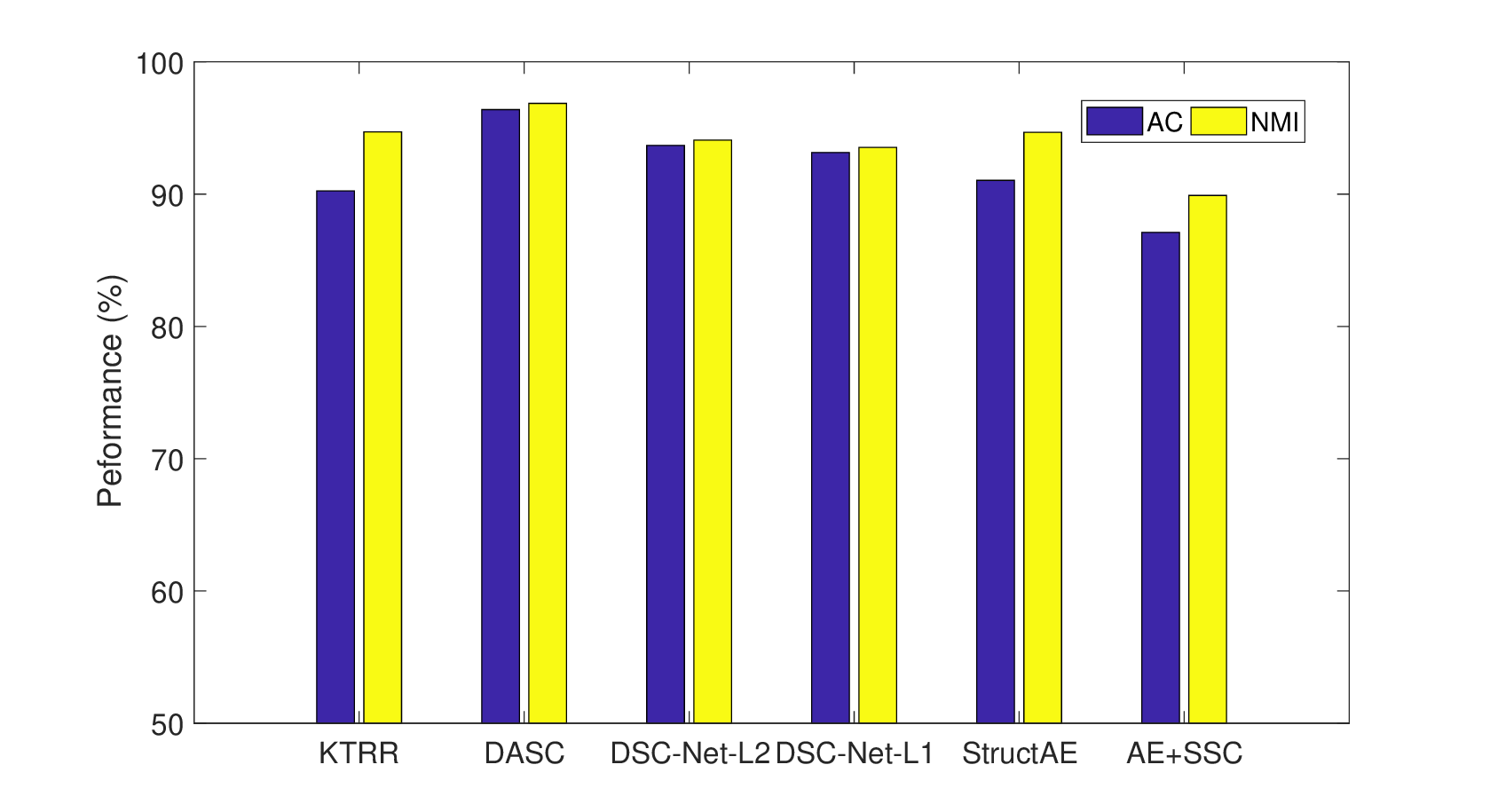}
\caption{The clustering performance comparison between our method and the deep learning-based clustering methods on the COIL20 dataset.}
\label{fig:deepcom}
\end{figure}

The deep learning methods have achieved great success in numerous recognition tasks. In this experiment, we compare the proposed methods with deep subspace cluster methods. The comparison result\footnote{The results of DASC and StructAE are presented by their authors.} on the COIL20 dataset is shown in Fig.~\ref{fig:deepcom}, from which we can see that KTRR outperforms AE+SSC in terms of Accuracy and NMI. Even KTRR is inferior to DASC~\cite{zhou2018deep}, DSC-Net-L2~\cite{Ji2017deep}, DSC-Net-L1~\cite{Ji2017deep} and StructAE~\cite{peng2018stru} in terms of Accuracy, it achieves competitive scores as these four deep subspace clustering methods in terms of NMI. Note that deep learning-based methods perform well only when sufficiently large amounts of data are available. This is a severe limitation in fields in which obtaining such data is either difficult or expensive. KTRR does not need a large number of data samples, and it has a closed-form solution, which costs much less computational resources to obtain the final clustering results.

\subsection{Comparison of Performance on Large-Scale Database}
To evaluate the capability of handling large-scale data, we compare seven peer algorithms on the Covtype database. Since the cost of computing the values of ARI and F-Score on the large-scale database is extremely high, we only report Accuracy, NMI, and the time cost of the tested algorithms in Table \ref{covtyperesult}. From the results, we have the following observations:
\begin{itemize}

\item The extension of KTRR, EKTRR, outperforms the other algorithms. The result of Accuracy of EKTRR is $11.01\%$ higher than the second best algorithm SSSC, and $15.34\%$ higher than the second fastest algorithm Nystr\"{o}m.

\item SLSR obtains the highest NMI value, but the NMI values obtained by all the tested algorithms are very small. The metric NMI is not able to distinguish the performance of the tested algorithms in this case.

\item The time cost of our method is the lowest. It is at least two times faster than other algorithms. This is due to the fact that our method uses the in-sample data to train the neural network. Then, all other data are classified by the trained network. This makes the proposed method very competitive to handle large-scale databases. Some other

\end{itemize}

\begin{table}[htbp]\scriptsize
\centering
\caption{Clustering performance (mean $\pm$ sd, \%) and mean time cost (seconds) comparisons of different methods on the Covtype database (581,012 samples). The best mean results in different metrics are in bold. The ``$\dagger$'' indicates that the value of the proposed method is significantly different from all other methods at a $0.05$ level by the Wilcoxon's rank sum test.}
\begin{tabular}{l||lll}
\hline
Methods & AC          & NMI         & Time            \\
\hline
\hline
EKTRR & \textbf{39.61}$\pm$\textbf{5.91}$\dagger$ &5.19$\pm$ 2.71$\dagger$& \textbf{16.50}$\dagger$\\
SSSC~\cite{Peng2015b} & 28.60$\pm$0.00 & 5.30 $\pm$ 0.00 & 135.62 \\
SLSR~\cite{Peng2015b} &26.45$\pm$0.00 & \textbf{7.14}$\pm$\textbf{0.00} & 119.78 \\
SLRR~\cite{Peng2015b} &27.23$\pm$0.03 & 3.65$\pm$0.02 & 122.85 \\
KASP~\cite{Yan2009FAS} &23.95$\pm$1.96 & 3.55$\pm$0.18 & 913.25 \\
Nystr\"{o}m~\cite{Chen2011parallel} &24.26$\pm$0.61 & 3.75$\pm$0.04 & 22.95 \\
SEC~\cite{nie2011spectral} &21.05$\pm$0.01 & 3.64$\pm$0.01 & 32.05 \\
AKK~\cite{Chitta2011AKK} &22.76$\pm$1.65 & 3.76$\pm$0.08 & 240.65 \\
\hline
\end{tabular}
\label{covtyperesult}
\end{table}

\section{Conclusion}
\label{Sec:5}
In this paper, we incorporated the kernel technique into the linear representation method to achieve robust nonlinear subspace clustering. It does not need the prior knowledge about the structure of errors in the input data and remedies the drawback of the existing TRR method that it cannot deal with the data points from nonlinear subspaces. Moreover, through the theoretical analysis of our proposed mathematical model, we found that the developed optimisation problem can be solved analytically, and the closed-form solution is only dependent on the kernel matrix. These advantages enable our proposed method being potentially useful in many real-world applications. Comprehensive experiments on several real-world image databases have demonstrated the effectiveness and efficiency of the proposed method.

In the future, we plan to conduct a systematical investigation on the selection of optimal kernel for our proposed method and study how to determine the number of nonlinear subspaces automatically.


\bibliographystyle{IEEEtran}
\begin{small}
\bibliography{mybib}
\end{small}

\end{document}